\documentclass[sigconf]{acmart}

\renewcommand\footnotetextcopyrightpermission[1]{}
\usepackage{booktabs}

\usepackage{bm}
\usepackage[mathscr]{euscript}
\usepackage{algorithm}
\usepackage{algpseudocode}
\usepackage{color}
\usepackage[inline]{enumitem}
\usepackage{tikz}
\usetikzlibrary{bayesnet}

\newcommand{\WRP}{\par\qquad\(\hookrightarrow\)\enspace}

\setcopyright{rightsretained}

\graphicspath{{./graphics/}}

\begin{document}
\title{Introduction to Tensor Decompositions and their\\Applications in Machine Learning}

\author{Stephan Rabanser}
\affiliation{
	\department{Department of Informatics}
  	\institution{Technical University of Munich}
}
\email{rabanser@in.tum.de}

\author{Oleksandr Shchur}
\affiliation{
	\department{Department of Informatics}
  	\institution{Technical University of Munich}
}
\email{shchur@in.tum.de}

\author{Stephan G\"unnemann}
\affiliation{
	\department{Department of Informatics}
  	\institution{Technical University of Munich}
}
\email{guennemann@in.tum.de}

\begin{abstract}
Tensors are multidimensional arrays of numerical values and therefore generalize matrices to multiple dimensions. While tensors first emerged in the psychometrics community in the $20^{\text{th}}$ century, they have since then spread to numerous other disciplines, including machine learning. Tensors and their decompositions are especially beneficial in unsupervised learning settings, but are gaining popularity in other sub-disciplines like temporal and multi-relational data analysis, too. 

The scope of this paper is to give a broad overview of tensors, their decompositions, and how they are used in machine learning. As part of this, we are going to introduce basic tensor concepts, discuss why tensors can be considered more rigid than matrices with respect to the uniqueness of their decomposition, explain the most important factorization algorithms and their properties, provide concrete examples of tensor decomposition applications in machine learning, conduct a case study on tensor-based estimation of mixture models, talk about the current state of research, and provide references to available software libraries.
\end{abstract}

\maketitle

\section{Introduction}

Tensors are generalizations of matrices to higher dimensions and can consequently be treated as multidimensional fields. 

Tensors and their decompositions originally appeared in 1927 \cite{hitchcock:tensor}, but have remained untouched by the computer science community until the late $20^{\text{th}}$ century \cite{sidiropoulos:tensor}. Fueled by increasing computing capacity and a better understanding of multilinear algebra especially during the last decade, tensors have since expanded to other domains, like statistics, data science, and machine learning \cite{kolda:tensor}. 

In this paper, we will first motivate the use of and need for tensors through Spearman's hypothesis and evaluate low-rank matrix decomposition approaches, while also considering the issues that come with them. We will then introduce basic tensor concepts and notation, which will lay the groundwork for the upcoming sections. In particular, we will analyze why low-rank tensor decompositions are much more rigid compared to low-rank matrix decompositions. Then, we will turn to some of the most widely used tensor decompositions, CP and Tucker, and the theory behind them, while also elaborating on their most important properties and key differences. Also, we will explain how tensor decompositions help us with uncovering underlying hidden low-dimensional structure in the tensor. Finally, we will explain why and how tensors and their decomposition can be used to tackle typical machine learning problems and afterwards look into two concrete examples of a tensor-based parameter estimation method for spherical Gaussian mixture models (GMMs) and single topic models. By using the proposed method, we can extract all needed information from low-order moments of the underlying probability distribution to learn simple GMMs and topic models in an efficient way. We will close by highlighting available tensor software libraries and by presenting the most prominent open research questions in the tensor field and recap some of the key learnings.

\section{Matrix Decomposition: A Motivating Example}
\label{sec:matrix_decomp}

Matrix decompositions are important techniques used in different mathematical settings, such as the implementation of numerically efficient algorithms, the solution of linear equation systems, and the extraction of quintessential information from a matrix. As part of this section, we will focus on the \textit{rank decomposition} of a matrix, an information extraction technique, which can be formally expressed as
\begin{equation}
	\bm{M} = \bm{A}\bm{B}^{T} \quad \text{with}\quad \bm{M} \in \mathbb{R}^{n \times m},\ \bm{A} \in \mathbb{R}^{n \times r},\ \bm{B}^{T} \in \mathbb{R}^{r \times m}
\end{equation}
where $r$ represents the rank of the decomposition. Intuitively, this decomposition aims at explaining the matrix $\bm{M}$ through $r$ different latent factors, which are encoded in the matrices $\bm{A}$ and $\bm{B}^{T}$.

The problem with lots of matrix factorization approaches is the fact that they are considered non-unique, meaning that a number of different matrices $\bm{A}$ and $\bm{B}^{T}$ can give rise to a specific $\bm{M}$ \cite{moitra:tensor, moitra:tensor2}. In order to ensure uniqueness, additional constraints need to be imposed on a matrix, like positive-definiteness or orthogonality. In contrast, tensors do not require such strong constraints in order to offer a unique decomposition. They provide much better identifiability conditions through the usage of higher dimensions, as we will see in Section \ref{sec:tensor_intro}.
 
Before starting the discussion on tensors, we will briefly introduce Spearman's hypothesis and the rotation problem as an example motivating the use of tensors.

\subsection{Spearman's Hypothesis}

\begin{figure}
\includegraphics[width=1.0\linewidth]{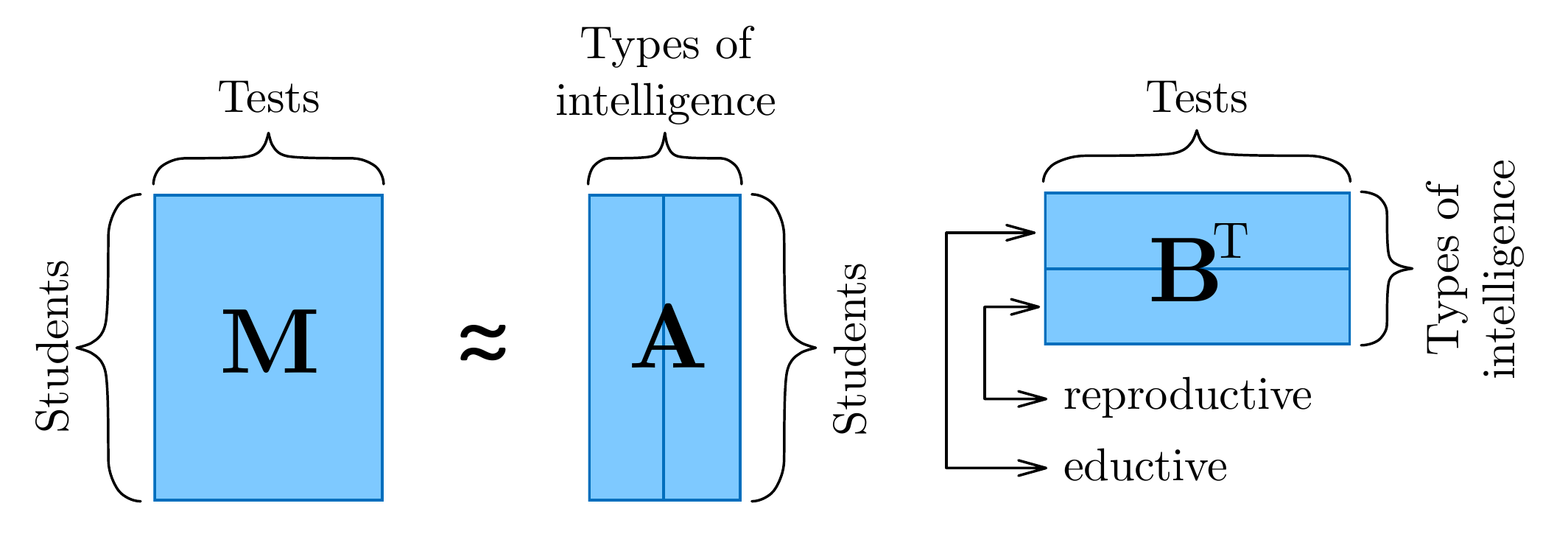}
\caption{Spearman's hypothesis}
\label{fig:spearman}
\end{figure}

In 1904, Charles Spearman, a British psychologist, supposed that human intelligence can be broken down into two (hidden) factors: \textit{eductive} (the ability to make sense out of complexity) and \textit{reproductive} (the ability to store and reproduce information) intelligence \cite{jensen1985nature}. Spearman therefore was one of the first scientists to carry out an unsupervised learning technique on a data set, which is nowadays referred to as \textit{factor analysis}.

To test his hypothesis, he invited $s$ students to take $t$ different tests and noted down their scores in a matrix $\bm{M} \in \mathbb{R}^{s \times t}$. He was then wondering whether it would be possible to uniquely decompose the matrix $\bm{M}$ into two smaller matrices $\bm{A} \in \mathbb{R}^{s \times h}$ and $\bm{B}^T \in \mathbb{R}^{h \times t}$ through the $h = 2$ latent factors described above. According to his explanation, $\bm{A}$ would hold a list of students with their corresponding eductive and reproductive capabilities and $\bm{B}^T$ would hold a list of tests with their corresponding eductive and reproductive requirements. This problem setting is depicted in Figure~\ref{fig:spearman}.

\subsection{The Rotation Problem}
\label{sec:rotation}

Given a matrix $\bm{M}$, we would like to approximate it as well as possible with another matrix $\hat{\bm{M}}$ of a lower rank (for Spearman's hypothesis: rank($\hat{\bm{M}}$) = $2$). 
Formally, the objective can be defined as minimizing the norm of the difference between the two matrices:
\begin{equation} \label{eq:matrix_decomp}
	\min_{\hat{\bm{M}}} ||\bm{M} - \hat{\bm{M}}||\quad \text{with}\quad \hat{\bm{M}} = \bm{A}\bm{B}^T
\end{equation}
\begin{figure}
\includegraphics[width=1.0\linewidth]{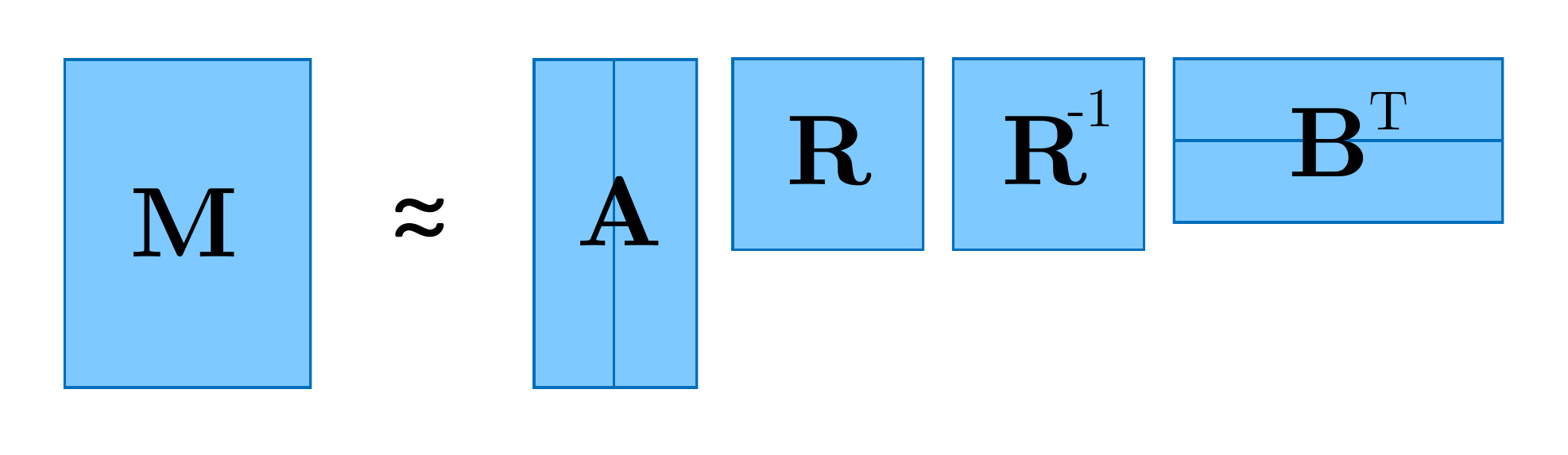}
\caption{The rotation problem}
\label{fig:rotation}
\end{figure}
However, this decomposition is not unique. By inserting an invertible rotation matrix $\bm{R}$ together with its inverse $\bm{R}^{-1}$ between $\bm{A}$ and $\bm{B}^T$ and absorbing $\bm{R}$ on the left with $\bm{A}$ and $\bm{R}^{-1}$ on the right with $\bm{B}^T$ we can again construct two matrices $\tilde{\bm{A}}$ and $\tilde{\bm{B}}^T$ \cite{moitra:tensor, moitra:tensor2}. This problem is usually referred to as the \textit{rotation problem} and is depicted in Figure \ref{fig:rotation}.
\begin{equation}
	\begin{split}
		\hat{\bm{M}} = \bm{A}\bm{B}^T = \bm{A}\bm{R}\bm{R}^{-1}\bm{B}^T & = (\bm{A}\bm{R})(\bm{R}^{-1}\bm{B}^T)\\ & = (\bm{A}\bm{R})(\bm{B}\bm{R}^{-T})^T = \tilde{\bm{A}}\tilde{\bm{B}}^T
	\end{split}
\end{equation}
We have seen that the rank decomposition of a matrix is generally highly non-unique. We conclude that matrix decompositions are only unique under very stringent conditions, such as orthogonality constraints which are imposed by the singular value decomposition (SVD) \cite{moitra:tensor}. Soon we will see that tensor decompositions are usually unique under much milder conditions.

\section{Introduction to Tensors}
\label{sec:tensor_intro}

\subsection{Basics}

As we have already learned, tensors can be thought of as multi-way collections of numbers, which typically come from a field (like $\mathbb{R}$). In the simplest high-dimensional case, such a tensor would be a three-dimensional array, which can be thought of as a data cube. Throughout this paper, we will often refer to a three-dimensional tensor for motivation and simplicity. In most cases, the notation naturally extends to higher-dimensional tensors. As we introduce different concepts in this and the next sections, we will borrow most of our notation from the comprehensive reviews of Kolda et al. \cite{kolda:tensor} and Sidiropoulos et al. \cite{sidiropoulos:tensor}.

\subsubsection{Tensor Order}

\begin{figure}
\includegraphics[width=1.0\linewidth]{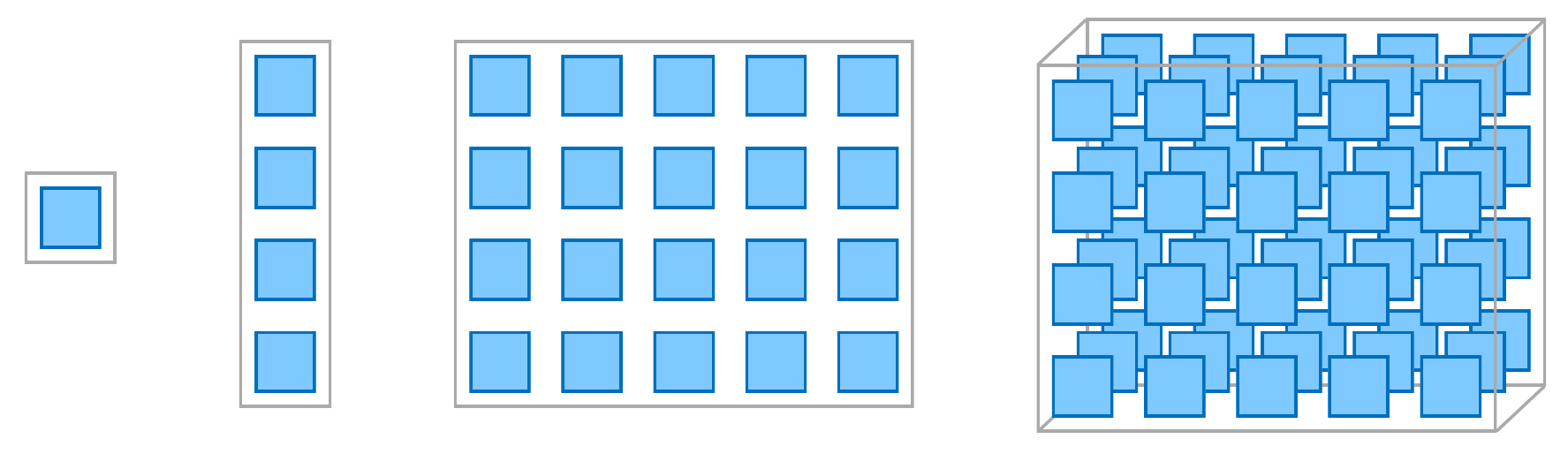}
\caption{$x \in \mathbb{R}$, $\bm{x} \in \mathbb{R}^{4}$, $\bm{X} \in \mathbb{R}^{4 \times 5}$, $\bm{\mathscr{X}} \in \mathbb{R}^{4 \times 5 \times 3}$}
\label{fig:data_store}
\end{figure}

The \textit{order}\footnote{The order of a tensor is sometimes also referred to as its \textit{way} or \textit{mode}.} of a tensor is the number of its dimensions. Scalars can therefore be interpreted as zeroth-order tensors, vectors as first-order tensors, and matrices as second-order tensors. We will refer to tensors of order three or higher as higher-order tensors. Notation-wise, scalars are denoted by lower case letters $x \in \mathbb{R}$, vectors by lower case bold letters $\bm{x} \in \mathbb{R}^{I_1}$, matrices by upper case bold letters $\bm{X} \in \mathbb{R}^{I_1 \times I_2}$, and higher order tensors by upper case bold Euler script letters $\bm{\mathscr{X}} \in \mathbb{R}^{I_1 \times I_2 \times \ldots \times I_N}$. The $I$s denote the number of elements in the respective dimension. Figure \ref{fig:data_store} shows how we can move from scalars to tensors.

\subsubsection{Tensor Indexing}

\begin{figure}
\includegraphics[width=1.0\linewidth]{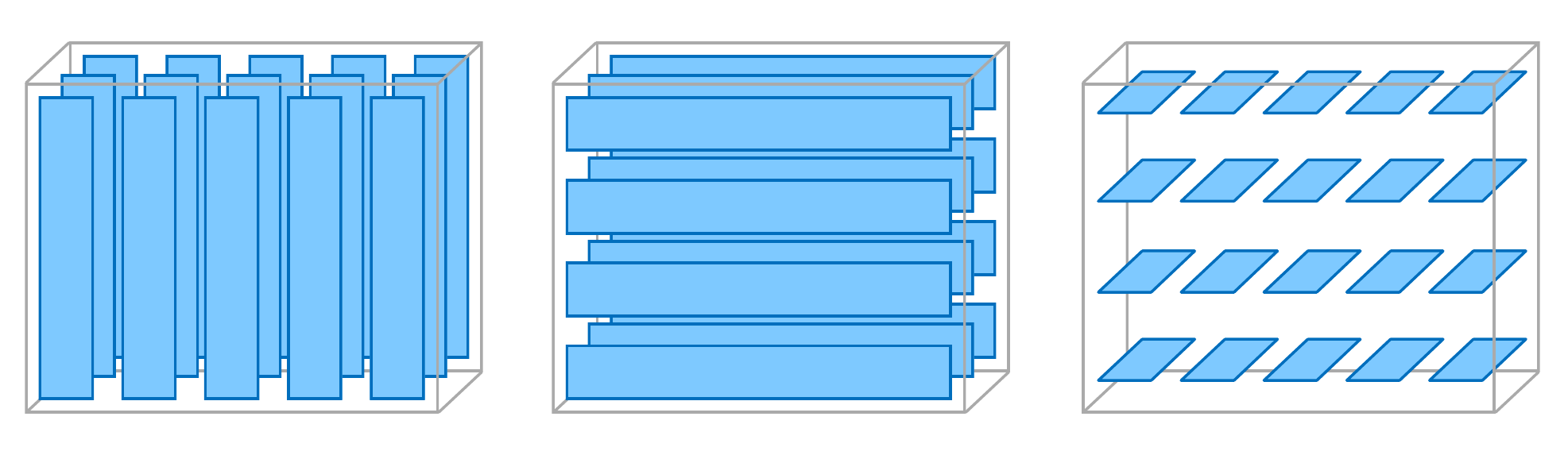}
\caption{Column, row, and tube fibers of a mode-3 tensor}
\label{fig:fibers}
\end{figure}

\begin{figure}
\includegraphics[width=1.0\linewidth]{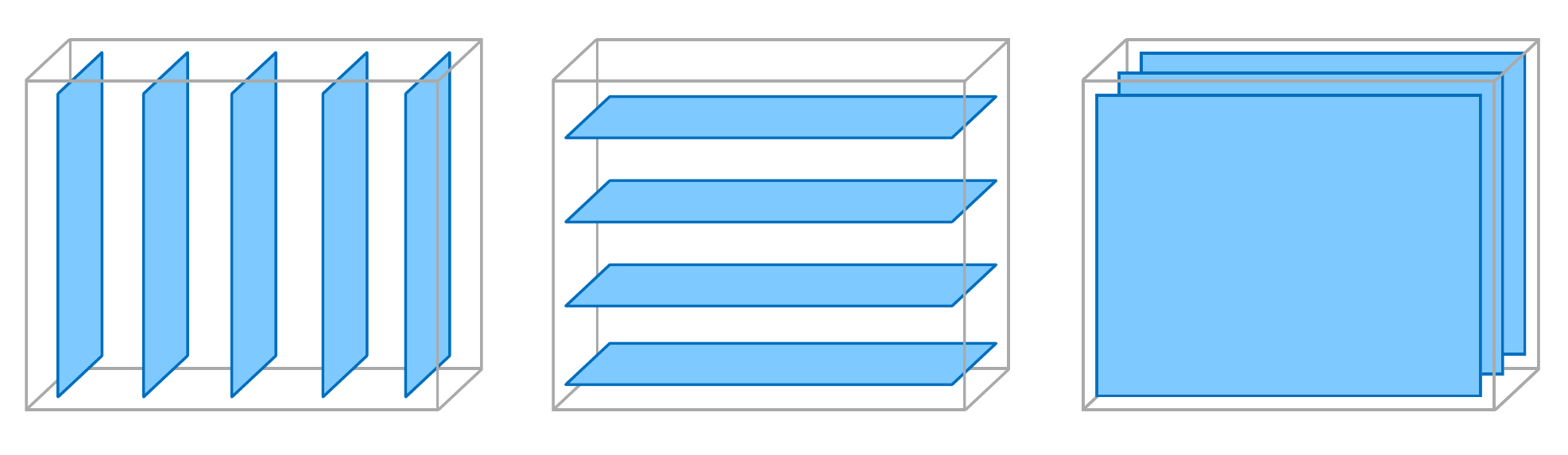}
\caption{Lateral, horizontal, and frontal slices of a mode-3 tensor}
\label{fig:slices}
\end{figure}

We can create subarrays (or subfields) by fixing some of the given tensor's indices. \textit{Fibers} are created when fixing all but one index, \textit{slices} (or slabs) are created when fixing all but two indices. For a third order tensor the fibers are given as $\bm{x}_{:jk} = \bm{x}_{jk}$ (column), $\bm{x}_{i:k}$ (row), and $\bm{x}_{ij:}$ (tube); the slices are given as $\bm{X}_{::k} = \bm{X}_{k}$ (frontal), $\bm{X}_{:j:}$ (lateral), $\bm{X}_{i::}$ (horizontal). Graphical examples of fibers and slices for a 3-way tensor are given in Figure~\ref{fig:fibers} and \ref{fig:slices}.

\subsubsection{Outer and Inner Product}

The \textit{vector outer product} is defined as the product of the vector's elements. This operation is denoted by the $\circledcirc$ symbol\footnote{Some publications denote the tensor product with the $\otimes$ symbol which we will use to denote the Kronecker product.}. The vector outer product of two $n$-sized vectors $\bm{a}$, $\bm{b}$ is defined as follows and produces a matrix $\bm{X}$:
\begin{equation}
	\bm{X} = \bm{a} \circledcirc \bm{b} = \bm{a} \bm{b}^T
\end{equation}
By extending the vector outer product concept to the general \textit{tensor product} for $N$ vectors, we can produce a tensor $\bm{\mathscr{X}}$: 
\begin{equation}
	\label{eq:rank_one}
  \bm{\mathscr{X}} = \bm{a}^{(1)} \circledcirc \bm{a}^{(2)} \circledcirc \cdots \circledcirc \bm{a}^{(N)}\ \text{with}\ x_{i_1i_2\cdots i_N} = a_{i_1}^{(1)} a_{i_2}^{(2)} \cdots a_{i_N}^{(N)}
\end{equation}
In contrast to the outer product, the \textit{inner product} of two $n$-sized vectors $\bm{a}$, $\bm{b}$ is defined as
\begin{equation}
	x = {\displaystyle \langle \bm{a}, \bm{b} \rangle = \bm{a}^T \bm{b} = \sum _{i=1}^{n}a_{i}b_{i}=a_{1}b_{1}+a_{2}b_{2}+\cdots +a_{n}b_{n}}
\end{equation}
and produces a scalar $x$.

\subsubsection{Rank-1 Tensors}

\begin{figure}
\includegraphics[width=0.75\linewidth]{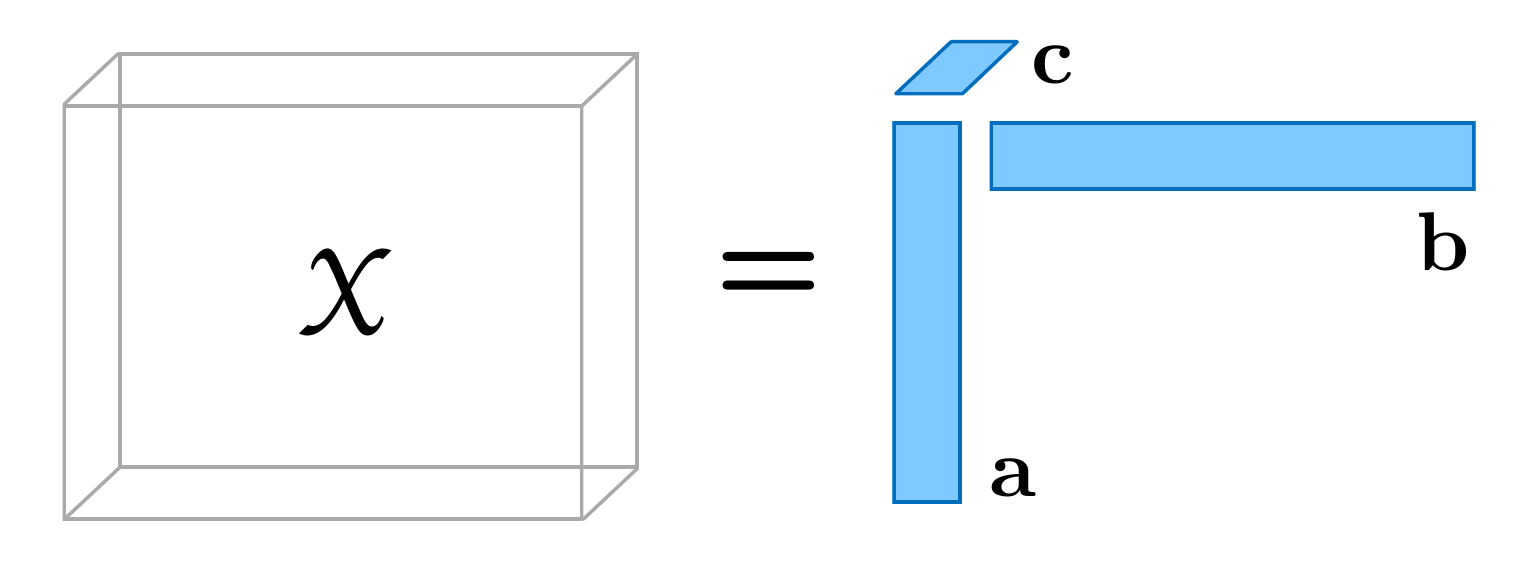}
\caption{A rank-1 mode-3 tensor}
\label{fig:rank_one}
\end{figure}

A $N$-way tensor is of \textit{rank-1} if it can be strictly decomposed into the outer product of $N$ vectors. Intuitively, this means that we introduce different scalings of a sub-tensor as we add more dimensions when building up the complete tensor. A rank-one matrix can therefore be written as $\bm{X} = \bm{a} \circledcirc \bm{b}$ and a rank-one 3-way tensor as $\bm{\mathscr{X}} = \bm{a} \circledcirc \bm{b} \circledcirc \bm{c}$. The general N-way form was already introduced in Equation \eqref{eq:rank_one}. A graphical view of the rank-1 concept is given in Figure \ref{fig:rank_one}.

\subsubsection{Tensor Rank}

The \textit{rank} of a tensor $\text{rank}(\bm{\mathscr{X}}) = R$ is defined as the minimum number of rank-one tensors which are needed to produce $\bm{\mathscr{X}}$ as their sum. A rank-$R$ matrix can therefore be written as $\bm{X} = \sum_{r=1}^{R} \lambda_r \bm{a}_r \circledcirc \bm{b}_r = \llbracket \bm{\lambda}; \bm{A}, \bm{B} \rrbracket$ and a rank-$R$ 3-way tensor as $\bm{\mathscr{X}} = \sum_{r=1}^{R} \lambda_r \bm{a}_r \circledcirc \bm{b}_r \circledcirc \bm{c}_r = \llbracket \bm{\lambda}; \bm{A}, \bm{B}, \bm{C} \rrbracket$. The general N-way form is given as:
\begin{equation}
	\boxed{
	\begin{split}
		\bm{\mathscr{X}} & = \sum_{r=1}^{R} \lambda_r \bm{a}_r^{(1)} \circledcirc \bm{a}_r^{(2)} \circledcirc \cdots \circledcirc \bm{a}_r^{(N)} \\ & =  \llbracket \bm{\lambda}; \bm{A}^{(1)}, \bm{A}^{(2)}, \cdots, \bm{A}^{(N)} \rrbracket
	\end{split}
  	}
\end{equation}
The $\bm{A}$s are called \textit{factor matrices} and hold the combination of the vectors from the rank-one components as columns. A specific $\bm{A}$ therefore has the form $\bm{A} = \begin{bmatrix} \bm{a}_1 & \bm{a}_2 & \cdots & \bm{a}_R \end{bmatrix}$.

We introduced a new additional factor $\lambda_r$ which is often used to absorb the respective weights during normalization of the factor matrices' columns. This usually means normalizing the sum of the squares of the elements in each column to one. Note that $\bm{\lambda} \in \mathbb{R}^{R}$. This notation will be especially useful once turn to machine learning applications of tensor decompositions in Section \ref{sec:tensor_ml} and once we introduce the Tucker decomposition in Section \ref{sec:tucker}. 

\subsection{Tensor Reorderings}

\subsubsection{Vectorization}

We can turn a given matrix $\bm{X} \in \mathbb{R}^{I \times J}$ into a vector by vertically stacking the columns of $\bm{X}$ into a tall vector.
\begin{equation}
	\text{vec}(\bm{X}) = \begin{bmatrix} 
			\bm{x}_{:1}\\
			\bm{x}_{:2}\\
			\vdots\\
			\bm{x}_{:J}
		\end{bmatrix}
\end{equation}

\subsubsection{Matricization}

Analogously, \textit{matricization} is the operation that reorders a tensor into a matrix. While there are other ways of rearranging vector elements into a matrix, we will only look into the mode-$n$ matricization of a tensor. The mode-$n$ matricization (or unfolding) of a tensor $\bm{\mathscr{X}} \in \mathbb{R}^{I_1 \times I_2 \times \ldots \times I_N}$, denoted $\bm{X}_{(n)} \in \mathbb{R}^{I_{n} \times (I_{1} \cdot \ldots \cdot I_{n-1} \cdot I_{n+1} \cdot \ldots \cdot I_{N})}$, turns the mode-$n$ fibers of $\bm{\mathscr{X}}$ into the columns of $\bm{X}_{(n)}$.

Let $x \in \bm{\mathscr{X}}$ be an element of a tensor and $m \in \bm{M}$ be an element of the unfolded tensor. Then we can define the mode-$n$ matricization via the following mapping:
\begin{equation}
	x_{i_1,i_2,\cdots,i_N} \mapsto m_{i_n,j}\ \text{with}\ j = 1 + \sum_{\substack{k=1\\k \neq n}}^{N}\bigg((i_k - 1)\prod_{\substack{m=1\\m \neq n}}^{k-1}I_m \bigg)
\end{equation}
To illustrate the formula introduced above, we will present a brief example of the matricization of a third-order tensor from \cite{kolda:tensor}. Let $\bm{\mathscr{X}}$ be a tensor with the following frontal slices:
		\begin{equation*}
			\small
			\bm{X}_1 = \begin{bmatrix} 
					1 & 4 & 7 & 10 \\
					2 & 5 & 8 & 11 \\
					3 & 6 & 9 & 12 \\
				\end{bmatrix} \qquad \bm{X}_2 = \begin{bmatrix} 
					13 & 16 & 19 & 22 \\
					14 & 17 & 20 & 23 \\
					15 & 18 & 21 & 24 \\
				\end{bmatrix}
		\end{equation*}
		Then the three mode-n matricizations are:
		\begin{equation*}
			\bm{X}_{(1)} = \begin{bmatrix} 
					1 & 4 & 7 & 10 & 13 & 16 & 19 & 22 \\
					2 & 5 & 8 & 11 & 14 & 17 & 20 & 23 \\
					3 & 6 & 9 & 12 & 15 & 18 & 21 & 24 \\
				\end{bmatrix}
		\end{equation*}
		\begin{equation*}
			\bm{X}_{(2)} = \begin{bmatrix} 
					1 & 2 & 3 & 13 & 14 & 15 \\
					4 & 5 & 6 & 16 & 17 & 18 \\
					7 & 8 & 9 & 19 & 20 & 21 \\
					10 & 11 & 12 & 22 & 23 & 24 \\
				\end{bmatrix}
		\end{equation*}
		\begin{equation*}
			\bm{X}_{(3)} = \begin{bmatrix} 
					1 & 2 & 3 & 4 & \cdots & 9 & 10 & 11 & 12 \\
					13 & 14 & 15 & 16 & \cdots & 21 & 22 & 23 & 12 \\
				\end{bmatrix}
		\end{equation*}

\subsection{Important Matrix/Tensor Products}

\subsubsection{Kronecker Product}

The \textit{Kronecker product} between two arbitrarily-sized matrices $\bm{A} \in \mathbb{R}^{I \times J}$ and $\bm{B} \in \mathbb{R}^{K \times L}$, $\bm{A} \otimes \bm{B} \in \mathbb{R}^{(IK) \times (JL)}$, is a generalization of the outer product from vectors to matrices:\begin{equation}
	\begin{split}
		\bm{A} \otimes \bm{B} & := \begin{bmatrix} 
			a_{11}\bm{B} & a_{12}\bm{B} & \cdots & a_{1J}\bm{B} \\
			a_{21}\bm{B} & a_{22}\bm{B} & \cdots & a_{2J}\bm{B} \\
			\vdots & \vdots & \ddots & \vdots \\
			a_{I1}\bm{B} & a_{I2}\bm{B} & \cdots & a_{IJ}\bm{B} \\
		\end{bmatrix}\\
		& = \begin{bmatrix} \bm{a}_1 \otimes \bm{b}_1 & \bm{a}_1 \otimes \bm{b}_2 & \cdots & \bm{a}_J \otimes \bm{b}_{L-1} & \bm{a}_J \otimes \bm{b}_{L} \end{bmatrix}
	\end{split}
\end{equation}

\subsubsection{Khatri-Rao Product}

The \textit{Khatri-Rao product} between two matrices $\bm{A} \in \mathbb{R}^{I \times K}$ and $\bm{B} \in \mathbb{R}^{J \times K}$, $\bm{A} \odot \bm{B} \in \mathbb{R}^{(IJ) \times K}$, corresponds to the column-wise Kronecker product.
\begin{equation}
	\bm{A} \odot \bm{B} := \begin{bmatrix} \bm{a}_1 \otimes \bm{b}_1 & \bm{a}_2 \otimes \bm{b}_2 & \cdots & \bm{a}_K \otimes \bm{b}_K \end{bmatrix}
\end{equation}

\subsubsection{Hadamard Product}

The \textit{Hadamard product} between two same-sized matrices $\bm{A} \in \mathbb{R}^{I \times J}$ and $\bm{B} \in \mathbb{R}^{I \times J}$, $\bm{A} * \bm{B} \in \mathbb{R}^{I \times J}$, corresponds to the element-wise matrix product.
\begin{equation}
	\bm{A} * \bm{B} := \begin{bmatrix} 
		a_{11}b_{11} & a_{12}b_{12} & \cdots & a_{1J}b_{1J} \\
		a_{21}b_{21} & a_{22}b_{22} & \cdots & a_{2J}b_{2J} \\
		\vdots & \vdots & \ddots & \vdots \\
		a_{I1}b_{I1} & a_{I2}b_{I2} & \cdots & a_{IJ}b_{IJ} \\
	\end{bmatrix}
\end{equation}

\subsubsection{$n$-mode Product}

The \textit{$n$-mode product} of a tensor $\bm{\mathscr{X}} \in \mathbb{R}^{I_1 \times I_2 \times \ldots \times I_N}$ and a matrix $\bm{M} \in \mathbb{R}^{J \times I_n}$ is denoted by $\bm{\mathscr{Y}} = \bm{\mathscr{X}} \times_n \bm{M}$ with $\bm{\mathscr{Y}} \in \mathbb{R}^{I_1 \times \cdots \times I_{n-1} \times J \times I_{n+1} \times \cdots \times I_N}$. Intuitively, each mode-$n$ fiber is multiplied by the matrix $\bm{M}$. The $n$-mode product can also be expressed through matricized tensors as $\bm{Y}_{(n)} = \bm{M}\bm{X}_{(n)}$. Element-wise, this operation can be expressed as follows:
\begin{equation}
	(\bm{\mathscr{X}} \times_n \bm{M})_{i_1\cdots i_{n-1}ji_{n+1}\cdots i_N} = \sum_{i_n = 1}^{I_n} x_{i_1\cdots i_N}m_{ji_n}
\end{equation}
The $n$-mode product also exists for tensors and vectors. The $n$-mode product of a tensor and a vector $\bm{v} \in \mathbb{R}^{I_n}$ is denoted by $\bm{\mathscr{Y}} = \bm{\mathscr{X}} \times_n \bm{v}$. Intuitively, each mode-$n$ fiber is multiplied by the vector $\bm{v}$. Element-wise, this operation can be expressed as follows:
\begin{equation}
	(\bm{\mathscr{X}} \times_n \bm{v})_{i_1\cdots i_{n-1}i_{n+1}\cdots i_N} = \sum_{i_n = 1}^{I_n} x_{i_1\cdots i_N}v_{i_n}
\end{equation}

\subsubsection{Multilinear Tensor Transformation} 

A tensor $\bm{\mathscr{X}}$ can be transformed on multiple dimensions by \textit{hitting} each of the vectors producing the tensor with a transformation matrix (or vector) from the left side \cite{anandkumar:tensor1}. Hitting a tensor $\bm{\mathscr{X}}$ with a matrix $\bm{M}_i$ on the $i$-th dimension corresponds to a matrix-vector multiplication between $\bm{M}_i$ and the vector on the $i$-th dimension, $\bm{M}_i^T \bm{a}^{(i)}$, thereby ensuring that the result of each individual transformation will again result in a vector. Instead, hitting a tensor $\bm{\mathscr{X}}$ with a vector $\bm{v}_i$ on the $i$-th dimension corresponds to an inner product between $\bm{v}_i$ and the vector on the $i$-th dimension, $\langle \bm{v}_i, \bm{a}^{(i)}\rangle$, resulting in a scalar.

In the 3-dimensional case, the multilinear tensor transformation using matrices is defined as follows:
\begin{equation}
	\small
	\boxed{\bm{\tilde{\mathscr{X}}} = \bm{\mathscr{X}}(\bm{M}_1,\bm{M}_2,\bm{M}_3) = \sum_{r=1}^{R} \lambda_r (\bm{M}_1^T \bm{a}_r) \circledcirc (\bm{M}_2^T \bm{b}_r) \circledcirc (\bm{M}_3^T \bm{c}_r)}
	\label{eq:mult_trans}
\end{equation}
This equation will prove valuable when we turn to tensor-based estimation of mixture models in Section \ref{sec:est_mm}.

\subsection{Tensor Uniqueness and Rigidness}

A tensor decomposition is called \textit{unique} if there exists only one combination of rank-1 tensors that sum to $\bm{\mathscr{X}}$ up to a common scaling and/or permutation indeterminacy. Intuitively, this means that there is one and only one decomposition and we can not construct a different arrangement of rank-1 tensors that sum to $\bm{\mathscr{X}}$. As we will see below, tensor decompositions are unique under much more relaxed requirements on the tensor compared to matrices and hence are considered more rigid than matrices.

Since we are usually interested in a low-rank tensor decomposition of $\bm{\mathscr{X}}$, let us now take a look at an interesting property of low-rank tensors. Given a low-rank tensor $\bm{\mathscr{X}}$, then any slice through that tensor 
\begin{equation}
	\bm{X}_{k} = \sum_{r=1}^{R} ( \bm{a}_r \circledcirc \bm{b}_r ) c_{kr}
\end{equation}
is in itself a low-rank matrix again. A low-rank tensor is therefore not just a collection of low-rank matrices, but there exist interrelations between these slices. We can easily observe, that all of these slices also share the same column and row spaces. Looking more closely, we can make an even stronger claim, namely that the slices are different scalings of the same set of rank-1 matrices \cite{moitra:tensor, moitra:tensor2}. This is a very strong constraint on the tensor's structure, which could help us address the rotation problem we described in Section \ref{sec:rotation}.

The low rank assumption enables us to determine whether the factors we found capture the underlying (latent) structure of the tensor. To do that we can subtract off scalings of the same rank-1 matrix, $\bm{X}_{k} - c_{k} ( \bm{a} \circledcirc \bm{b} )$, to decrease the rank of each slice of the tensor \cite{moitra:tensor, moitra:tensor2}. For matrices, there are many possible low-rank matrices one could use to strictly reduce its rank, but for tensors it is required that the same low-rank matrix works for all tensor slices. This strong interconnection between their slices makes tensors way more \textit{rigid} than matrices and allows for weaker uniqueness conditions.

\section{Tensor Decomposition Algorithms}
\label{sec:tensor_decomp}

After familiarizing ourselves with the basics of tensors, we will now turn to the most popular tensor decomposition algorithms. While this section will provide a rather theoretical treatment, we also describe the practical applicability of tensor decompositions in Section \ref{sec:tensor_ml} and Section \ref{sec:est_mm}.

In particular, we are wondering whether we can generalize the concept of the SVD from matrices to general tensors. As we will see, there is no single generalization of the SVD concept, but we will discuss two decompositions that feature different generalized properties of the matrix SVD: the \textit{canonical polyadic decomposition} (CPD) and the \textit{Tucker decomposition}. Both are outer product decompositions, but they have very different structural properties. As a rule of thumb it is usually advised to use CPD for latent parameter estimation and Tucker for subspace estimation, compression, and dimensionality reduction.

Since CPD and Tucker are the most important tensor decomposition and many other decompositions are based on these two techniques, discussing any other factorization approaches would go beyond the scope of this paper. Just like in Section \ref{sec:tensor_intro}, we will introduce these decompositions from Sidiropoulos et al. \cite{sidiropoulos:tensor} and Kolda et al. \cite{kolda:tensor}, which also provide deeper theoretical insights and the most important adaptions of the mentioned decompositions.

\subsection{Canonical Polyadic Decomposition (CPD)}

\begin{figure}
\includegraphics[width=1.0\linewidth]{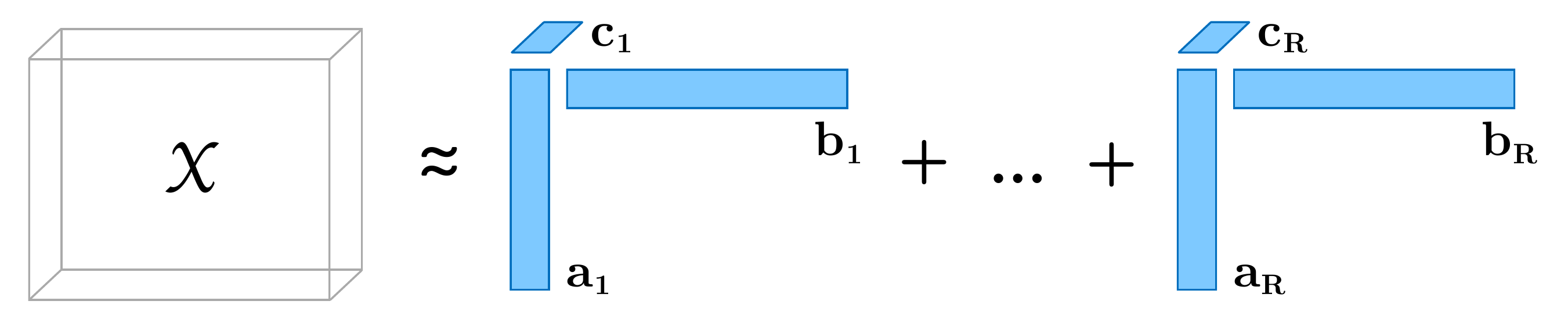}
\caption{CP Decomposition}
\label{fig:cp_decomp}
\end{figure}

The first type of tensor decomposition we will consider are rank decompositions. The key concept of rank decompositions is to express a tensor as the sum of a finite number of rank-one tensors. The most prominent rank decompositions are the CANonical DECOMPosition (CANDECOMP) and the PARAllel FACtors (PARAFAC) decomposition. Both have their origins in different knowledge domains and have been independently discovered many times, but they both boil down to the same principles, which is why we will from here on refer to this type of decomposition as the CANDECOMP/PARAFAC or canonical polyadic decomposition (CPD).

We can formalize the 3-way CPD case as follows:
\begin{equation}
	\label{eq:cpd}
	\boxed{
	\min_{\hat{\bm{\mathscr{X}}}} ||\bm{\mathscr{X}} - \hat{\bm{\mathscr{X}}}||\ \ \ \text{where}\ \ \ \hat{\bm{\mathscr{X}}} = \sum_{r=1}^{R} \bm{a}_r \circledcirc \bm{b}_r \circledcirc \bm{c}_r = \llbracket \bm{A}, \bm{B},\bm{C} \rrbracket
	}
\end{equation}

A graphical view of this concept is given in Figure \ref{fig:cp_decomp}.

Note the very similar structure to the matrix decomposition problem in \eqref{eq:matrix_decomp}. In the exact case, meaning that if $\min_{\hat{\bm{\mathscr{X}}}} ||\bm{\mathscr{X}} - \hat{\bm{\mathscr{X}}}|| = 0$, we refer to $\hat{\bm{\mathscr{X}}}$ being an exact low rank approximation to $\bm{\mathscr{X}}$. We can restate this problem in a matricized form, too:
\begin{equation}
	\begin{split}
		\hat{\bm{X}}_{(1)} & = (\bm{C} \odot \bm{B})\bm{A}^T\\
		\hat{\bm{X}}_{(2)} & = (\bm{C} \odot \bm{A})\bm{B}^T\\
		\hat{\bm{X}}_{(3)} & = (\bm{B} \odot \bm{A})\bm{C}^T\\
	\end{split}
\end{equation}
For the general case we have:
\begin{equation}
	\hat{\bm{\mathscr{X}}} = \sum_{r=1}^{R} \lambda_r \bm{a}_r^{(1)} \circledcirc \bm{a}_r^{(2)} \circledcirc \cdots \circledcirc \bm{a}_r^{(n)} = \llbracket \bm{\lambda}; \bm{A}^{(1)}, \bm{A}^{(2)}, \cdots, \bm{A}^{(n)} \rrbracket
\end{equation}
\begin{equation}
	\hat{\bm{X}}_{(n)} = \bm{\Lambda}(\bm{A}^{(N)} \odot \cdots \odot \bm{A}^{(n+1)} \odot \bm{A}^{(n-1)} \odot \cdots \odot \bm{A}^{(1)})\bm{A}^{(n)T}\\
\end{equation}
Note that $\bm{\Lambda} = \text{Diag}(\bm{\lambda})$.

There exist different algorithms to compute CPD of a given tensor and we will take a brief look at two of the most popular ones.

\subsubsection{Jennrich's Algorithm}
	
If $\bm{A}$, $\bm{B}$, and $\bm{C}$ are all linearly independent (i.e. have full rank), then $\bm{\mathscr{X}} = \sum_{r=1}^{R} \lambda_r \bm{a}_r \circledcirc \bm{b}_r \circledcirc \bm{c}_r$ is unique up to trivial rank permutation and feature scaling and we can use Jennrich's algorithm to recover the factor matrices \cite{moitra:tensor, moitra:tensor2}. The algorithm works as follows:
\begin{enumerate}
	\item Choose random vectors $\bm{x}$ and $\bm{y}$.
	\item Take a slice through the tensor by hitting the tensor with the random vector $\bm{x}$:\\$\bm{\mathscr{X}}(\bm{I}, \bm{I}, \bm{x}) = \sum_{r = 1}^{R} \langle \bm{c}_r, \bm{x} \rangle \bm{a}_r \circledcirc \bm{b}_r = \bm{A} \text{Diag}(\langle \bm{c}_r, \bm{x} \rangle) \bm{B}^T$.
	\item Take a second slice through the tensor by hitting the tensor with the random vector $\bm{y}$:\\$\bm{\mathscr{X}}(\bm{I}, \bm{I}, \bm{y}) = \sum_{r = 1}^{R} \langle \bm{c}_r, \bm{y} \rangle \bm{a}_r \circledcirc \bm{b}_r = \bm{A} \text{Diag}(\langle \bm{c}_r, \bm{y} \rangle) \bm{B}^T$.
	\item Compute eigendecomposition to find $\bm{A}$: \\ $\bm{\mathscr{X}}(\bm{I}, \bm{I}, \bm{x})\ \bm{\mathscr{X}}(\bm{I}, \bm{I}, \bm{y})^{\dagger} = \bm{A} \text{Diag}(\langle \bm{c}_r, \bm{x} \rangle) \text{Diag}(\langle \bm{c}_r, \bm{y} \rangle)^{\dagger} \bm{A}^{\dagger}$
	\item Compute eigendecomposition to find $\bm{B}$: \\ $\bm{\mathscr{X}}(\bm{I}, \bm{I}, \bm{x})^{\dagger}\ \bm{\mathscr{X}}(\bm{I}, \bm{I}, \bm{y}) = (\bm{B}^{T})^{\dagger} \text{Diag}(\langle \bm{c}_r, \bm{x} \rangle)^{\dagger} \text{Diag}(\langle \bm{c}_r, \bm{y} \rangle) \bm{B}^{T}$
	\item Pair up the factors and solve a linear system to find $\bm{C}$. 
\end{enumerate}
While this algorithm works well for some problems, it only takes random slices of a tensor and hence does not use the full tensor structure. Moreover, it requires good eigen-gap\footnote{difference between two successive eigenvalues} on the eigendecompositions of the factor matrices, the lack of which could lead to numerical instability \cite{hsu:latent}.

\subsubsection{Alternating Least Squares (ALS) Algorithm}

An other way of computing the CP decomposition of a tensor, and in fact the work-horse of modern tensor decomposition approaches, is the \textit{alternating least squares} (ALS) algorithm. The key idea behind this algorithm is to fix all factor matrices except for one in order to optimize for the non-fixed matrix and then repeat this step for every matrix repeatedly until some stopping criterion is satisfied.

For the 3-way tensor case, the ALS algorithm would perform the following steps repeatedly until convergence.
\begin{equation}
	\boxed{
	\begin{split}
		\bm{A} & \leftarrow \arg \min_{\bm{A}}||\bm{X}_{(1)} - (\bm{C} \odot \bm{B})\bm{A}^T||\\
		\bm{B} & \leftarrow \arg \min_{\bm{B}}||\bm{X}_{(2)} - (\bm{C} \odot \bm{A})\bm{B}^T||\\
		\bm{C} & \leftarrow \arg \min_{\bm{C}}||\bm{X}_{(3)} - (\bm{B} \odot \bm{A})\bm{C}^T||\\
	\end{split}
	}
\end{equation}
The optimal solution to this minimization problem is given by
\begin{equation}
	\begin{split}
		\hat{\bm{A}} & = \bm{X}_{(1)} [(\bm{C} \odot \bm{B})^{T}]^{\dagger} = \bm{X}_{(1)} (\bm{C} \odot \bm{B})(C^{T}C * B^{T}B)^{\dagger}\\
		\hat{\bm{B}} & = \bm{X}_{(2)} [(\bm{C} \odot \bm{A})^{T}]^{\dagger} = \bm{X}_{(2)} (\bm{C} \odot \bm{A})(C^{T}C * A^{T}A)^{\dagger}\\
		\hat{\bm{C}} & = \bm{X}_{(3)} [(\bm{B} \odot \bm{A})^{T}]^{\dagger} = \bm{X}_{(3)} (\bm{B} \odot \bm{A})(B^{T}B * A^{T}A)^{\dagger}\\
	\end{split}
\end{equation}
The generalization to the order-N case is given below. Note that due to the problem definition in Equation \eqref{eq:cpd}, the ALS algorithm requires the rank which should be used for the approximation as an argument \cite{kolda:tensor}.
\begin{algorithm}
\begin{algorithmic}
\small
\Procedure{CP-ALS}{$\bm{\mathscr{X}}$, R}
	\State initialize $\bm{A}^{(n)} \in R^{I_n \times R}\ \text{for}\ n = 1,\ldots,N$
	\Repeat
		\For{n = 1,\ldots,N} 
		\State $\bm{V} \leftarrow \bm{A}^{(1)T}\bm{A}^{(1)} * \cdots * \bm{A}^{(n-1)T}\bm{A}^{(n-1)} * \bm{A}^{(n+1)T}\bm{A}^{(n+1)} * $\WRP$\cdots * \bm{A}^{(N)T}\bm{A}^{(N)}$
		\State $\bm{A}^{(n)} \leftarrow \bm{X}_{(n)}(\bm{A}^{(N)} \odot \cdots \odot \bm{A}^{(n+1)} \odot \bm{A}^{(n-1)} \odot \cdots \odot \bm{A}^{(1)})\bm{V}^{\dagger}$
		\State \textit{normalize columns of $\bm{A}^{(n)}$ (optional)}
		\State \textit{store norms as $\bm{\lambda}$ (optional)}
		\EndFor
	\Until{stopping criterion satisfied}
	\State \Return $\bm{\lambda}, \bm{A}^{(1)}, \ldots, \bm{A}^{(N)}$
\EndProcedure
\end{algorithmic}
\caption{ALS algorithm}
\label{alg:als}
\end{algorithm}

While the ALS algorithm outlined in Algorithm \ref{alg:als} is simple to understand and implement, it might take several steps to converge and it might also not converge to a global optimum. This means that the performance of this algorithm is influenced by its initialization.

\subsubsection{Tensor Power Method}
\label{sec:tpm}

The last CPD algorithm we will consider is the so-called \textit{tensor power method}, which we can use in the special case of all identical factor matrices and if the $\bm{a}_r$s are all orthogonal. This restricts the tensor we want to decompose to the following structure in the 3-way case: $\bm{\mathscr{X}} = \sum_{r=1}^{R} \lambda_r \bm{a}_r \circledcirc \bm{a}_r \circledcirc \bm{a}_r$ \cite{anandkumar2014}. Note that the tensor power method is an analogue to the matrix power method. While the latter aims at finding the top singular vectors in a matrix, the former tries to determine the top singular vectors in a tensor.

The core idea of the matrix power method is described by the following recurrence relation, which computes an approximation $\bm{a}_{r,i+1}$ to the eigenvector $\bm{a}_{r}$ corresponding to the most dominant eigenvalue $\lambda_r$ based on $\bm{X}$ \cite{panju2011}:
\begin{equation}
	\label{eq:matrix_power}
	\bm{a}_{r,i+1} = \frac{\bm{X}_r(\bm{I},\bm{a}_{r,i})}{||\bm{X}_r(\bm{I},\bm{a}_{r,i})||_2} = \frac{\bm{X}_r\bm{a}_{r,i}}{||\bm{X}_r\bm{a}_{r,i}||_2}
\end{equation}
This approximation exploits the eigenvalue/-vector relationship $\bm{X}\bm{a}_r = \bm{X}(\bm{I},\bm{a}_r) = \lambda_r\bm{a}_r$. Note that the division by the $L_2$-normalized matrix-vector product allows us to get a better feeling for the convergence and that the initial vector $\bm{a}_{r,0}$ can be picked randomly or can be initialized with some correlation to the true eigenvector (if such knowledge is available) \cite{panju2011}. The top singular value $\lambda_r$ can be computed from $\bm{a}_{r,i}$ using normalization, formally $\lambda_r = ||\bm{a}_{r,i}||_2$, after convergence.

As we might be interested in not just extracting the very first dominant eigenvalue/-vector combination ($r$ const.), but the first $k$ ($r \in [k]$) \footnote{$x \in [k] \equiv x \in \{0, 1, \ldots, k\}$} or all such combinations ($r \in [R]$), it is important to \textit{deflate} the matrix once the dominant eigenvalue/-vector combination has been found \cite{anandkumar:tensor1}. This means removing $\lambda_r$ and $\bm{a}_r$ from the matrix as follows:
\begin{equation}
	\bm{X}_{r + 1} = \bm{X}_r - \lambda_r \bm{a}_r \circledcirc \bm{a}_r
\end{equation}
In order to transfer this concept to tensors, we only need to perform a few small adjustments. The two crucial equations are given as
\begin{equation}
	\boxed{
	\begin{split}		
		\bm{a}_{r,i+1} & = \frac{\bm{\mathscr{X}}_r(\bm{I},\bm{a}_{r,i},\bm{a}_{r,i})}{||\bm{\mathscr{X}}_r(\bm{I},\bm{a}_{r,i},\bm{a}_{r,i})||_2}\\
		\bm{\mathscr{X}}_{r + 1} & = \bm{\mathscr{X}}_r - \lambda_r \bm{a}_{r} \circledcirc \bm{a}_{r} \circledcirc \bm{a}_{r}
	\end{split}
	}
\end{equation}
where 
\begin{equation}
	\bm{\mathscr{X}}_r(\bm{I},\bm{a}_{r,i},\bm{a}_{r,i}) = \sum_{j \in [k]} \lambda_{r,j} \langle \bm{a}_{r,j}, \bm{a}_{r,i} \rangle^2 \bm{a}_{r,j} = \lambda_r \bm{a}_{r,i}
\end{equation}
which follows from Equation \eqref{eq:mult_trans} \cite{anandkumar:tensor1}. Section \ref{sec:dec_tpm} will give an example of how the tensor power method can be applied to a concrete machine learning problem.

\subsubsection{Uniqueness}

As we have seen in Section \ref{sec:rotation}, rank decompositions are (generally) not unique for matrices, whereas this is often the case for higher-order tensors. In order to further analyze this property, we first have to introduce the concept of $k$-rank. The \textit{$k$-rank} (or Kruskal rank) of a matrix $\bm{M}$, denoted $k_{\bm{M}}$, corresponds to the maximum number $k$ such that \textbf{any} $k$ columns are linearly independent. A sufficient uniqueness condition for the general CPD case would then be:
\begin{equation}
	\sum_{n=1}^{N}k_{\bm{A}^{(n)}} \geq 2R + (N-1)
\end{equation}
While the the above equation provides a sufficient condition for uniqueness, it is not necessary. Necessary conditions for uniqueness are:
\begin{equation}
	\min_{1,\ldots,N} \text{rank}(\bm{A}^{(1)} \odot \cdots \odot \bm{A}^{(n-1)} \odot \bm{A}^{(n+1)} \odot \cdots \odot \bm{A}^{(1)}) = R
\end{equation}
\begin{equation}
	\min_{1,\ldots,N} \bigg(\prod_{\substack{m=1\\m \neq n}}^{N} \text{rank}(\bm{A}^{(m)}) \bigg) \geq R
\end{equation}

\subsubsection{Tensor Rank Peculiarities}

Finally, since we are looking at a rank decomposition, we still want to shed some light on some tensor rank peculiarities. While the definition of tensor rank is a natural extension of the matrix rank, some of its properties are noticeably different.

\begin{itemize}
	\item There is no trivial algorithm to determine the rank of a tensor as the problem is NP-hard \cite{hillar:tensor}. Most algorithms therefore try to determine the fit for multiple CPDs and then pick the one which yields the best approximation. We note though that in practice a 100\% fit is close to impossible, as data is usually corrupted by noise and in fact the fit alone cannot determine the rank in such cases.
	\item The rank of a tensor with real entries can be different depending on the field it is defined over. For $\bm{\mathscr{X}} \in \mathbb{R}^{I_1 \times \ldots \times I_N}$ and $\hat{\bm{\mathscr{X}}} \in \mathbb{C}^{I_1 \times \ldots \times I_N}$ where $\bm{\mathscr{X}} = \hat{\bm{\mathscr{X}}}$ it is possible that $\text{rank}(\bm{\mathscr{X}}) \neq \text{rank}(\hat{\bm{\mathscr{X}}})$.
	\item The rank of a three-way tensor $\bm{\mathscr{X}} \in \mathbb{R}^{I \times J \times K}$ is bounded from above by the following inequality:
		\begin{equation}
			\text{rank}(\bm{\mathscr{X}}) \leq \min\{IJ,IK,JK\}
		\end{equation}
	\item The rank of a tensor is constant under mode permutation.
	\item While the best rank-$k$ approximation for a matrix is given by the truncated SVD, this is not true for tensors. In fact it is even possible that the best rank-$k$ approximation of a tensor does not exist at all. We call such tensors degenerate, which means that they can be approximated arbitrarily well by a lower-rank factorization.
	More formally, a tensor $\mathscr{X}$ of $\text{rank}(\bm{\mathscr{X}}) = m$ is called degenerate if
	\begin{equation}
	\forall \epsilon > 0 \quad \exists \bm{\mathscr{\hat{X}}}, \, \text{rank}(\bm{\mathscr{\hat{X}}}) = k < m \; \text{ s.t. } ||\bm{\mathscr{X}} - \hat{\bm{\mathscr{X}}}|| < \epsilon
	\end{equation}

\end{itemize}

\subsection{Tucker Decomposition}
\label{sec:tucker}

The Tucker decomposition decomposes a tensor into a so-called core tensor and multiple matrices which correspond to different core scalings along each mode. Therefore, the Tucker decomposition can be seen as a higher-order PCA.

In the 3-way tensor case, we can express the problem of finding the Tucker decomposition of a tensor $\bm{\mathscr{X}} \in \mathbb{R}^{I \times J \times K}$ with $\bm{\mathscr{G}} \in \mathbb{R}^{P \times Q \times R}$, $\bm{A} \in \mathbb{R}^{I \times P}$, $\bm{B} \in \mathbb{R}^{J \times Q}$, $\bm{C} \in \mathbb{R}^{K \times R}$ as follows:
\begin{equation}
	\boxed{
	\begin{split}
		\min_{\hat{\bm{\mathscr{X}}}} ||\bm{\mathscr{X}} - \hat{\bm{\mathscr{X}}}||\quad \text{with}\quad \hat{\bm{\mathscr{X}}} & = \sum_{p=1}^{P}\sum_{q=1}^{Q}\sum_{r=1}^{R} g_{pqr} \bm{a}_r \circledcirc \bm{b}_r \circledcirc \bm{c}_r\\
		& = \bm{\mathscr{G}} \times_1 \bm{A} \times_2 \bm{B} \times_3 \bm{C}\\
		& = \llbracket \bm{\mathscr{G}}; \bm{A}, \bm{B},\bm{C} \rrbracket
	\end{split}
	}
\end{equation}

\begin{figure}
\includegraphics[width=1.0\linewidth]{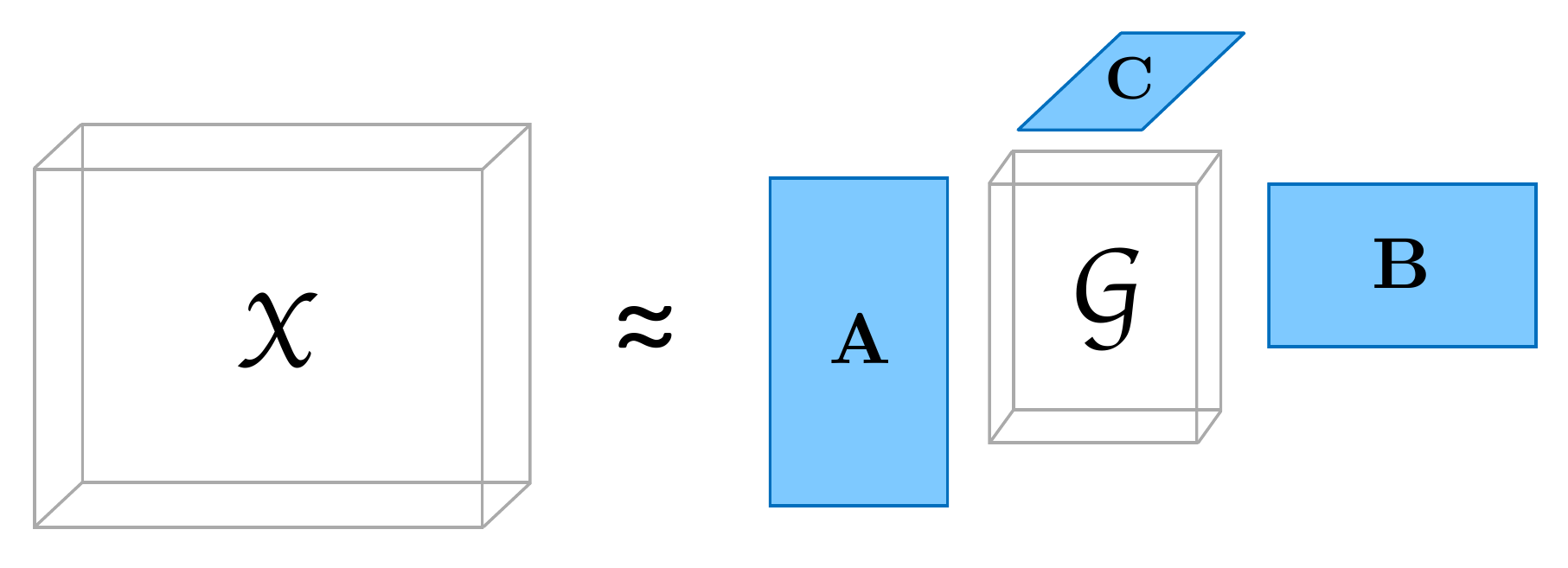}
\caption{Tucker Decomposition}
\label{fig:tucker_decomp}
\end{figure}

A graphical view of this concept is given in Figure \ref{fig:tucker_decomp}.

In this setting, $\bm{\mathscr{G}}$ is the core tensor, which expresses how and to which extend different tensor elements interact with each other. The factor matrices $\bm{A}$, $\bm{B}$, and $\bm{C}$ are often referred to as the principal component in the respective tensor mode. We can already see, that if we pick $P < I$, $Q < J$, and $R < K$, this will result in a compression of $\bm{\mathscr{X}}$, with $\bm{\mathscr{G}}$ being the compressed version of $\bm{\mathscr{X}}$.

The matricized version of the above tensor is given as:
\begin{equation}
	\begin{split}
		\hat{\bm{X}}_{(1)} = \bm{A}\bm{G}_{(1)}(\bm{C} \otimes \bm{B})^{T}\\
		\hat{\bm{X}}_{(2)} = \bm{B}\bm{G}_{(1)}(\bm{C} \otimes \bm{A})^{T}\\
		\hat{\bm{X}}_{(3)} = \bm{C}\bm{G}_{(1)}(\bm{B} \otimes \bm{A})^{T}\\
	\end{split}
\end{equation}
In the general N-way case we get:
\begin{equation}
	\begin{split}
		\hat{\bm{\mathscr{X}}} & = \sum_{r_1=1}^{R_1}\sum_{r_2=1}^{R_2}\cdots \sum_{r_N=1}^{R_N} g_{r_1r_2\cdots r_N} \bm{a}_{i_1r_1}^{(1)} \circledcirc \cdots \circledcirc \bm{a}_{i_Nr_N}^{(N)}\\
		& = \bm{\mathscr{G}} \times_1 \bm{A}^{(1)} \times_2 \cdots \times_N \bm{A}^{(N)}\\
		& = \llbracket \bm{\mathscr{G}}; \bm{A}^{(1)}, \ldots, \bm{A}^{(N)} \rrbracket
	\end{split}
\end{equation}
\begin{equation}
	\hat{\bm{X}}_{(n)} = \bm{A}^{(n)}\bm{G}_{(n)}(\bm{A}^{(N)} \otimes \cdots \otimes \bm{A}^{(n+1)} \otimes \bm{A}^{(n-1)} \otimes \cdots \otimes \bm{A}^{(1)})^{T}
\end{equation}

Before we can look into the exact computation of the Tucker decomposition, we first need to introduce the concept of $n$-rank. The $n$-rank of a tensor $\bm{\mathscr{X}} \in \mathbb{R}^{I_1 \times I_2 \times \ldots \times I_N}$ corresponds to the column rank of the $n$-th unfolding of the tensor $\bm{X}_{(n)}$. Formally, this is denoted $\text{rank}_n(\bm{\mathscr{X}})$. It is important to not confuse this concept with the tensor rank.

\subsubsection{Higher Order Singular Value Decomposition (HOSVD)}

We can now make use of the fact that for a given tensor $\bm{\mathscr{X}}$, we can easily find an exact Tucker decomposition of rank $(R_1,R_2,\ldots,R_N)$ where $R_n = \text{rank}_n(\bm{\mathscr{X}})$. This gives rise to the \textit{higher order singular value decomposition} (HOSVD). The key idea behind the HOSVD is to find the components that best capture the variation in mode $n$, while not considering the other modes at this point in time \cite{kolda:tensor}. This directly corresponds to the basic PCA concept and can be formalized as seen in Algorithm \ref{alg:hosvd}.

\begin{algorithm}
\begin{algorithmic}
\Procedure{HOSVD}{$\bm{\mathscr{X}}, R_1, \ldots, R_N$}
	\For{n = 1,\ldots,N} 
		\State $\bm{A}^{(n)} \leftarrow R_n$ leading left singular vectors of $\bm{X}_{(n)}$
	\EndFor
	\State $\bm{\mathscr{G}} \leftarrow \bm{\mathscr{X}} \times_1 \bm{A}^{(1)T} \times_2 \cdots \times_N \bm{A}^{(N)T}$
	\State \Return $\bm{\mathscr{G}}, \bm{A}^{(1)}, \ldots, \bm{A}^{(N)}$
\EndProcedure
\end{algorithmic}
\caption{HOSVD}
\label{alg:hosvd}
\end{algorithm}

\subsubsection{Higher Order Orthogonal Iteration (HOOI)}

An alternative approach to computing the Tucker decomposition is provided by the \textit{higher order orthogonal iteration} (HOOI). The HOOI is essentially an ALS algorithm, which uses the outcome of performing HOSVD on a tensor as a starting point for initializing the factor matrices \cite{kolda:tensor}. This algorithm, which is outlined in Algorithm \ref{alg:hooi}, is especially advised in cases where we only have access to a truncated HOSVD, since the successive application of the ALS algorithm allows for more accurate decompositions.

\begin{algorithm}
\begin{algorithmic}
\small
\Procedure{HOOI}{$\bm{\mathscr{X}}, R_1, \ldots, R_N$}
	\State initialize $\bm{A}^{(n)} \in R^{I_n \times R}\ \text{for}\ n = 1,\ldots,N$ using HOSVD
	\Repeat
		\For{n = 1,\ldots,N} 
		\State $\bm{\mathscr{Y}} \leftarrow \bm{\mathscr{X}} \times_1 \bm{A}^{(1)T} \times_2 \cdots \times_{n-1} \bm{A}^{(n-1)T} \times_{n+1} \bm{A}^{(n+1)T} \times_{n+2} $\WRP$ \cdots \times_{N} \bm{A}^{(N)T}$
		\State $\bm{A}^{(n)} \leftarrow R_n$ leading left singular vectors of $\bm{Y}_{(n)}$
		\EndFor
	\Until{stopping criterion satisfied}
	\State $\bm{\mathscr{G}} \leftarrow \bm{\mathscr{X}} \times_1 \bm{A}^{(1)T} \times_2 \cdots \times_N \bm{A}^{(N)T}$
	\State \Return $\bm{\mathscr{G}}, \bm{A}^{(1)}, \ldots, \bm{A}^{(N)}$
\EndProcedure
\end{algorithmic}
\caption{HOOI}
\label{alg:hooi}
\end{algorithm}

\subsubsection{Non-Uniqueness}

In contrast to the CPD and the (matrix) SVD, the Tucker decomposition is generally not unique. This intuitively follows from the fact that the core tensor $\bm{\mathscr{G}}$ can be arbitrarily structured and might allow interactions between any component. Imposing additional constraints on the structure of $\bm{\mathscr{G}}$ can therefore lead to more relaxed uniqueness properties. For instance, the CPD can be expressed in the Tucker model through a superdiagonal core tensor. The HOSVD generates an all-orthogonal core and hence relies on yet another type of special core structure.
\section{Tensor Applications in Machine Learning}
\label{sec:tensor_ml}

We will now briefly discuss how tensor decompositions can be used in various machine learning models and mention some example applications.

\subsection{Temporal Data}
Whenever some kind of relationship can be represented as a matrix (e.g. user preferences in a recommender system, adjacency matrix of a graph),
tensors provide a straightforward way to model the \textit{temporal component}.
Similarly to SVD and non-negative matrix factorization (NMF) in the case of a matrix, performing decomposition on the resulting tensor allows to detect latent structure in the data.
Typical tasks in temporal tensor analysis include discovering patterns \cite{xiong:temporal}, predicting evolution \cite{dunlavy:temporal} and spotting anomalies \cite{papalexakis:temporal}. 

As another example, the classic problem of community detection can be extended to finding so-called \textit{temporal communities},
that come into existence and subsequently disappear as time progresses \cite{araujo:temporal}.
Tensor methods can even be applied when new data arrives in a never-ending continuous stream, without having to deal with infinite time dimension \cite{sun:temporal}.

One important remark we would like to make is the fact that temporal data add jet another structural constraint to the tensor, restricting arbitrary permutations in terms of the tensor's dimensions. This results from the fact that the temporal interpretation adds an additional relationship between the data points stored in the tensor.

\subsection{Multi-relational Data} 

Another domain where tensors arise naturally is representation of \textit{multirelational data}.
As an example, one can think of social networks, where users communicate by exchanging messages, tagging pictures and following each other.
Each interaction here can be stored as a (subject, relation, object) triplet.
Such a multimodal structure is generalized by the notion of multilayer network \cite{kivela:multirel}.
Again, tensors lend themselves to a concise description of such data, with each slice representing one of the underlying relations.

Applying tensor factorization algorithms in this setting allows to determine interdependencies occurring on multiple levels simultaneously.
For instance, it enables to solve challenging tasks in \textit{statistical relational learning}, 
such as collective classification \cite{nickel:multirel1}, word representation learning \cite{jenatton:multirel},
community detection \cite{papalexakis:multirel}, and coherent subgraph learning \cite{boden2012mining}.

A prominent area of research concerned with multirelational data is analysis of \textit{knowledge networks}.
Knowledge graphs, such as Google Knowledge Graph, YAGO or Microsoft Academic Graph \cite{dong:knowledge},
are a special case of multilayer networks that are used to store facts about relationships between real-world entities.
Main challenge in analyzing such graphs is inferring new relations between objects given the existing data.
Here, tensor decomposition approaches provide state-of-the art performance in terms of both quality and computational efficiency \cite{nickel:multirel2, padia:multirel}.
This information can then be used in applications such as question answering and entity resolution \cite{socher:nlp}.

\subsection{Latent Variable Modeling}

One more area of machine learning where tensor decomposition methods have been gaining significant traction over the last decade is \textit{inference in latent variable models}.
Tensor methods have been successfully applied to hidden Markov models \cite{hsu:latent}, independent component analysis \cite{beckmann:latent} and topic models \cite{anandkumar:latent2}.

The standard problem setting is as follows: it is assumed that we are given a (generative) probabilistic model that describes how a set of hidden variables gives rise to observed data.
The inference problem now lies in determining the most likely setting of the hidden variables given the observations.
Classic algorithms such as maximum likelihood estimation are asymptotically consistent, but usually do not perform well in very high dimensions.

Tensor decomposition methods propose a different approach based on the so-called method of moments \cite{pearson:moments}.
The main idea lies in computing the empirical moments of the data (such as mean, variance, and skewness),
and then finding the configuration of latent variables that would give rise to similar quantities under the given model.
It has been shown in recent work that in many popular probabilistic models the low-order moment tensors exhibit a specific structure \cite{anandkumar:tensor1}.
This fact, combined with recent advances in multilinear algebra enables us to construct effective and efficient algorithms for solving such problems.
These methods are known to scale well to larger problems and in general do not suffer much from the curse of dimensionality \cite{oseledets:curse}.

We will provide more details on the method of moments, and show how it can be used to perform inference in Gaussian mixture model and topic model
by performing decomposition of the corresponding moment tensors.

\section{Case study: Estimation of mixture models}
\label{sec:est_mm}

In order to give the reader a tangible example of how tensor decompositions can be applied to a concrete machine learning problem, we will now take a more detailed look at how we can estimate the parameters for latent variable models by using tensor decompositions. While the basic concepts we present here will work for multiple different latent variable models, we will motivate the derivation with respect to two popular models in machine learning: a Gaussian mixture model and a topic model. Most of the following content is heavily based on Anandkumar et al. \cite{anandkumar:tensor1}, which also has additional insights on the computational efficiency and numerical aspects of the presented estimation procedure.

The goal of unsupervised learning approaches is to discover hidden structure (latent variables) in data where no labels are present during training. Typically, there are two main challenges in such settings.

\paragraph{Conditions for identifiability}
As one of the basic statistical questions we face in unsupervised learning problems, we have to determine whether a proposed model contains all relevant information for parameter derivation.

\paragraph{Efficient learning of latent variable models}
While maximum likelihood estimation (MLE) possesses nice properties such as asymptotic optimality and consistency, it is NP-hard in its derivation for a variety of different problem settings. In practice, iterative algorithms like expectation maximization (EM) are often used. Such algorithms usually suffer from slow convergence and local optima as they don't come with any consistency guarantees \cite{redner1984mixture}. In contrast, the presented way to learn GMMs and topic models is efficient, both with respect to computational complexity and statistical hardness barriers regarding the number of data points needed for learning.
\\[10px]
In the following, we will first introduce basic notation and concepts of two different mixture models, namely a Gaussian mixture model and a topic model. Next, we will explain the method of moments, which we will use to construct data moments up to order $3$. Finally, we will learn how to derive the wanted parameters from the third order moment by first whitening it and by consequently finding an eigendecomposition for this tensor through the tensor power method.

\subsection{Gaussian Mixture Model}

We will now briefly revise the concepts of Gaussian mixture modeling, which will be crucial for the upcoming paragraphs. A Gaussian mixture model is a probabilistic model which assumes that data points are generated from a mixture of $k$ different Gaussian distributions/clusters with unknown parameters. We will denote the hidden variable representing a specific Gaussian component by $\bm{h}$, which is a categorical variable and can take $k$ distinct states. With respect to the representation of $\bm{h}$, we will use the basis vector form, which means that $\bm{h} \in \{\bm{e}_1, \ldots, \bm{e}_k\}$ where $\bm{e}_i$ is the basis vector along the $i$-th direction. These vectors can therefore analogously be interpreted as 1-hot-encoded vectors. The expectation of $\bm{h}$ is defined as the probability of $\bm{h}$ taking one of these different states and is encoded on the vector $\bm{w}$: $\mathbb{E}[\bm{h}] = \bm{w}$. The means of the Gaussian mixture components are stored in the matrix $\bm{A} = [\bm{a}_1 \cdots \bm{a}_k]$ as columns where $\bm{a}_i$ is the mean of the $i$-th component. Hence, $\bm{A} \in \mathbb{R}^{d \times k}$ where $d$ corresponds to the dimensionality of the data. Each sample is generated as follows:
\begin{equation}
	\bm{x}_n = \bm{A}\bm{h} + \bm{z}\quad \text{with} \quad \bm{z} \sim \mathcal{N}(\bm{0}, \sigma^2\bm{I})
\end{equation}

\begin{figure}
\includegraphics[width=0.66\linewidth]{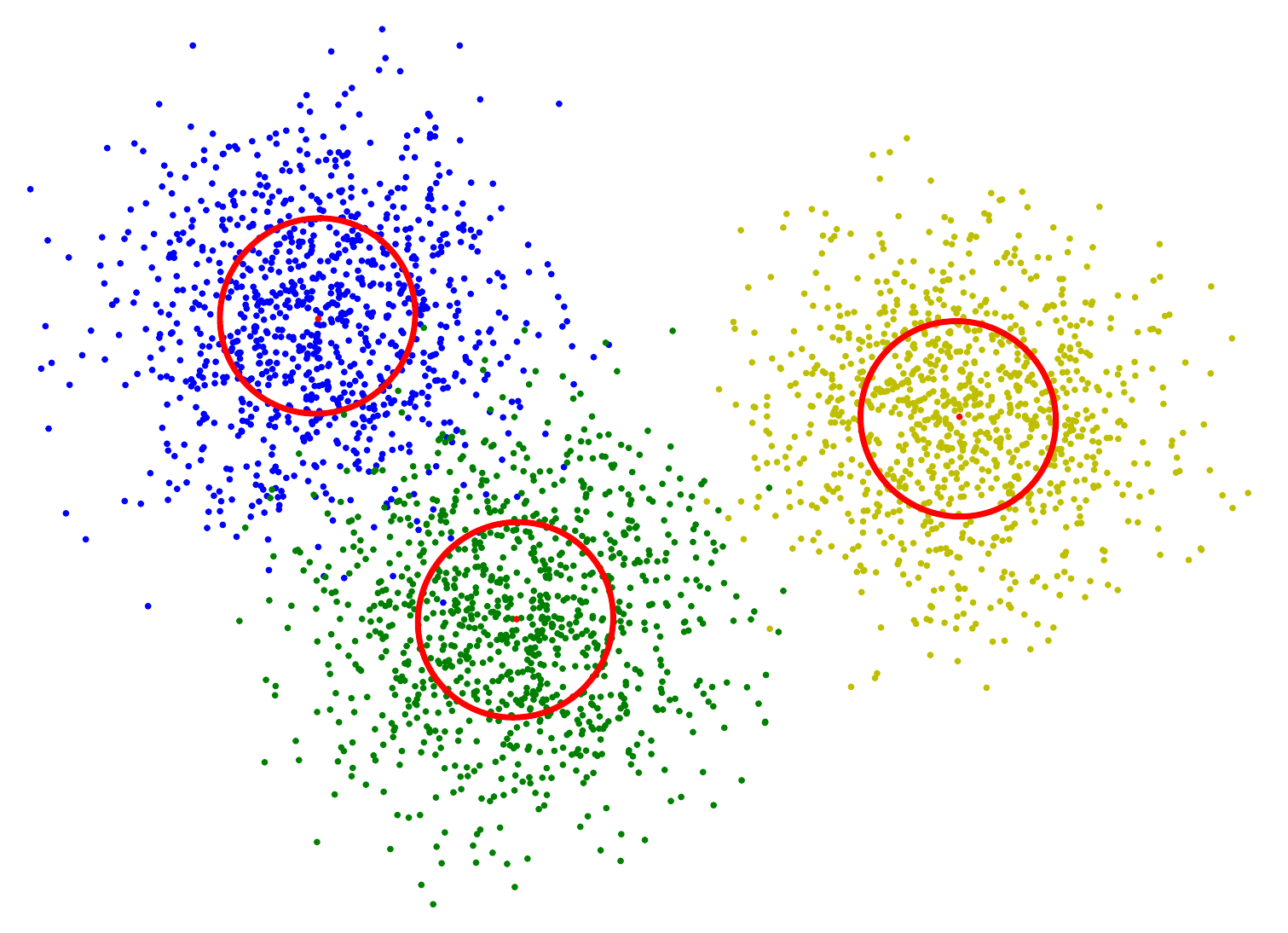}
\caption{Spherical Gaussian mixture example}
\label{fig:gmm}
\end{figure}

\begin{figure}
\begin{tikzpicture}
  \node[obs]                               	(xn) {$\bm{x}_n$};
  \node[latent, left=of xn, xshift=0cm] 		(A) {$\bm{A}$};
  \node[latent, above=of xn, xshift=0cm]		(h) {$\bm{h}$};
  \node[latent, right=1cm of xn]				(t) {$\bm{z}$};

  \edge {h,A,t} {xn};

  \plate {yx} {(h)(xn)} {$N$};
\end{tikzpicture}
\caption{Graphical model for Gaussian mixture model}
\label{fig:gm_gmm}
\end{figure}
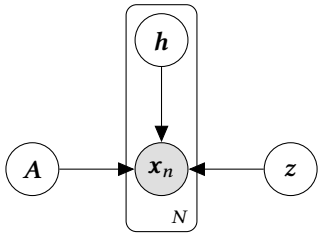

Note that we constrain the covariance of all Gaussian components to be spherical and identical for all components, as the covariance matrix is only governed by a single variance term $\sigma^2$ across all dimensions. While the generalization to differing but still spherical covariances is relatively easy to achieve, the generalization to arbitrary covariance matrices is more challenging and therefore beyond the scope of this paper \cite{ge:gmm}. 
Figure \ref{fig:gmm} shows samples drawn from a GMM with $k=3$ components and $d=2$ dimensions, while a graphical model is given in Figure \ref{fig:gm_gmm}.

\subsection{Topic Model}

\begin{figure}
\includegraphics[width=1.0\linewidth]{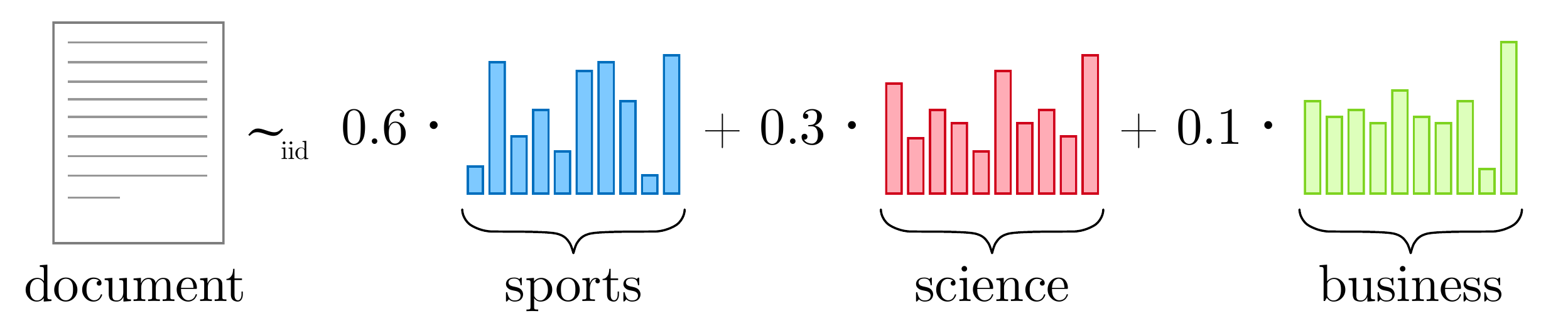}
\caption{General topic model distribution example}
\label{fig:topic_model}
\end{figure}

Another popular latent variable model is the so-called topic model, where we try to explain words in a document through hidden topics (see Figure~\ref{fig:topic_model}) in order to classify documents or to summarize them. A document consists of a mixture of $k$ different topics, again denoted by the hidden variable $\bm{h}$, from which $l \geq 3$ words are drawn from an exchangeable bag of words model with vocabulary dimension $d$ to form the text in the document. Every word therefore has the same probability of occurring in the text, given a specific topic. These probabilities are stored in a topic-word matrix $\bm{A} \in \mathbb{R}^{d \times k}$ with
\begin{equation}
	a_{ji} = P_{ji} = P(x_j | h_i)
\end{equation}
where the word $x_j$ is generated from topic $h_i$ with probability $P_{ji}$. 

\begin{figure}
\begin{tikzpicture}
  \node[obs, xshift=-2cm]     	(x1) {$\bm{x}_1$};
  \node[obs, xshift=-1cm]     	(x2) {$\bm{x}_2$};
  \node[obs, xshift=0cm]      	(x3) {$\bm{x}_3$};
  \node[obs, xshift=1cm]      	(d) {$\ldots$};
  \node[obs, xshift=2cm]      	(xn) {$\bm{x}_n$};
  \node[latent, above=of x3]		(h) {$\bm{h}$};

  \edge {h} {x1,x2,x3,d,xn};
\end{tikzpicture}
\caption{Graphical model for single topic model}
\label{fig:gm_stm}
\end{figure}
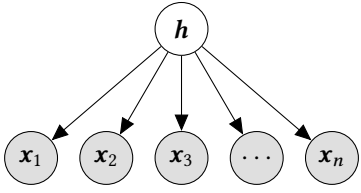

The most general treatment of the topic model setting is given by the latent Dirichlet allocation (LDA), where the probability distribution over the latent variable $\bm{h}$ is given by the Dirichlet distribution. Just as with the GMM, we will only take a look at a simplified model, where each document only contains one single topic, which is chosen with probability $w_i$ with $i \in [k]$. The latent variable $\bm{h} \in \{\bm{e}_1, \ldots, \bm{e}_k\}$ is therefore also again a categorical/1-hot-encoded variable and is therefore interpreted as the sole topic given a document. A graphical model is given in Figure \ref{fig:gm_stm}.

Since the vocabularies of different topics are discrete, the probability distributions over these vocabularies are discrete, too, in contrast to the GMM case. Therefore, the word distribution, given a certain topic, can also be thought of as a distribution over 1-hot-encoded vectors, which are of the size of the vocabulary. Our general linear model is consequently given as follows:
\begin{equation}
	\mathbb{E}_{\bm{x}_j \sim \bm{h}}[\bm{x}_j| \bm{h}] = \bm{A}\bm{h}
\end{equation}
Based on the topic-word matrix, the usage of 1-hot-encoded vectors for the words, and the fact that we assume $\bm{h}$ is fixed to a single topic $i$ per document, we can interpret the means as probability vectors:
\begin{equation}
	P(\bm{x}_j | \bm{t}_i) = \mathbb{E}_{\bm{x}_j \sim \bm{t}_i}[\bm{x}_j | \bm{t}_i] = \bm{a}_i
\end{equation}
Note that we write $\bm{t}_i$ (meaning "topic $i$") as an equivalent to $\bm{h} = \bm{e}_i \in \mathbb{R}^k$. Moreover, since we assume an exchangeable model where the words are sampled independently given a certain topic, the probability of a specific word conditioned on $\bm{h}$ is the same for all words.
\begin{equation}
	P(\bm{x} | \bm{t}_i) = P(\bm{x}_j | \bm{t}_i) \quad \forall j \in [d]
\end{equation}
This makes our subsequent moment calculations in the upcoming subsections a lot easier.

\subsection{Algorithm Overview}

Now that we are familiar with the basics of both models, we will give a brief overview of the parameter estimation procedure for both models.
\begin{enumerate}
	\item \textbf{Calculate moments}: Based on the method of moments, we can  find a moment formulation of the model in which the latent factors (the $\bm{a}_i$s and the weighting factors $w_i$) are producing these moments exclusively.
	\item \textbf{Whiten the data}: Once we have obtained the second and the third moment of the underlying data, we are still not able to extract the latent information from these two moments directly. Since the $\bm{a}_i$s producing these moments might not be orthogonal to each other, it is difficult to recover them uniquely for the moments. Therefore, we propose to orthogonalize (or whiten) the third moment to recover the $\bm{a}_i$s more easily.
	\item \textbf{Decompose the whitened moment tensor}: By using one of the available tensor decomposition methods, the tensor power method, we can then extract the latent factors $\bm{v}_i$ present in the whitened moment tensor.
	\item \textbf{Un-whiten the $\bm{v}_i$s}: As the resulting latent factors live in the whitened space, we have to transform them back to the original space by applying the inversion of the whitening transformaton.
\end{enumerate}

\subsection{Method of Moments (MoM)}
\label{sec:mom}

An alternative approach to classical MLE, which is often used for parameter estimation, is the previously discussed method of moments. The \textit{method of moments} tries to extract model information by building up higher order moments of the underlying probability distribution and therefore tries to infer model information based on averages of the data. Although we will not prove this fact, we claim that deriving first, second, and third order moment suffices to learn both the spherical GMM and the topic model. Most importantly, it is possible to reduce both (and even more) models to the following moment form:
\begin{equation}
	\label{eq:moments}
	\boxed{\bm{M}_2 = \sum_{i \in [k]} w_i\bm{a}_i \circledcirc \bm{a}_i \qquad \bm{\mathscr{M}}_3 = \sum_{i \in [k]} w_i\bm{a}_i \circledcirc \bm{a}_i \circledcirc \bm{a}_i}
\end{equation}
Recall that in the GMM case, the $\bm{a}_i$s correspond to the cluster mean vectors, whereas they represent word probability-vectors (given a certain topic) in the topic model case.

\subsubsection{GMM}

In the GMM setting, the first order moment is the mean of all Gaussian components which can be computed as follows:
\begin{equation}
	\bm{\mu} = \mathbb{E}[\bm{x}] = \bm{A}\bm{w} = \sum_{i \in [k]} w_i\bm{a}_i
\end{equation}
The  second moment corresponds to the covariance matrix of the mixture distribution and can be expressed as follows:
\begin{equation}
	\begin{split}
		\bm{\Sigma} = \mathbb{E}[\bm{x} \circledcirc \bm{x}] = \mathbb{E}[\bm{x}\bm{x}^T] & = \mathbb{E}[(\bm{A}\bm{h} + \bm{z})(\bm{A}\bm{h} + \bm{z})^T]\\
		& = \mathbb{E}[(\bm{A}\bm{h})(\bm{A}\bm{h})^T] + \mathbb{E}[\bm{z}\bm{z}^T]\\
		& = \sum_{i \in [k]} w_i(\bm{a}_i)(\bm{a}_i)^T + \sigma^2\bm{I}\\
		& = \sum_{i \in [k]} w_i\bm{a}_i \circledcirc \bm{a}_i + \sigma^2\bm{I}
	\end{split}
\end{equation}
By applying ($k-1$) principal component analysis (PCA)\footnote{Applying PCA with the rank of the projection dimension being $k-1$, i.e. $\min_{P \in \mathbb{R}^{d \times d}} \frac{1}{n	}\sum_{i \in [n]} ||x_i - Px_i||^2$ with $\text{rank}(P) = k-1$.} on the covariance matrix of this model, we can obtain $\text{span}(\bm{A})$. Also, we can already extract the common variance term $\sigma^2$ which corresponds to the smallest eigenvalue of the covariance matrix $\bm{\Sigma}$. If the covariances are governed by different variance terms, then the smallest eigenvalue will yield the average of all of these variances: $\overline{\sigma}^2 = \sum_{i=1}^{k}w_i \sigma_i^2$. Given these results, we can then apply a technique called spectral clustering, where we would project the samples onto $\text{span}(\bm{A})$ and try to classify the points there by using a distance-based clustering approach (e.g k-means). This method, however, requires that the clusters satisfy a sufficient separation criterion. In other words: the separation between the cluster means needs to be sufficiently large to cope with the variance in the data. Otherwise, too many points will be misclassified and hurt the model performance badly.

The third order moment, also called skewness\footnote{asymmetry of the probability distribution about its mean}, can be expressed in the following way:
\begin{equation}
	\begin{split}
		\bm{\mathscr{S}} = \mathbb{E}[\bm{x} \circledcirc \bm{x} \circledcirc \bm{x}] & = \mathbb{E}[(\bm{Ah}) \circledcirc (\bm{Ah}) \circledcirc (\bm{Ah})] \\ &\quad + \mathbb{E}[(\bm{Ah}) \circledcirc (\bm{z}) \circledcirc (\bm{z})] \\ &\quad + \mathbb{E}[(\bm{z}) \circledcirc (\bm{Ah}) \circledcirc (\bm{z})] \\ &\quad + \mathbb{E}[(\bm{z}) \circledcirc (\bm{z}) \circledcirc (\bm{Ah})]\\
		& = \sum_{i \in [k]} w_i \bm{a}_i \circledcirc \bm{a}_i \circledcirc \bm{a}_i + \sigma^2\sum_{i \in [d]}( \bm{\mu} \circledcirc \bm{e}_i \circledcirc \\ &\quad \circledcirc \bm{e}_i + \bm{e}_i \circledcirc \bm{\mu} \circledcirc \bm{e}_i + \bm{e}_i \circledcirc \bm{e}_i \circledcirc \bm{\mu})\\
	\end{split}
\end{equation}
Note that uneven-order moments of the underlying Gaussian components are $0$ and are therefore missing from these equation (uneven occurrence of $\bm{z}$ or $\bm{e}_i$ respectively).

At this point, both the shared variance $\sigma^2$ and the overall mean vector $\bm{\mu}$ are already known and we can simplify the equations for the second and third moment to bilinear and trilinear equations in the mean vectors, as expressed by Equation \eqref{eq:moments}.

\subsubsection{Topic model} In the (single) topic model setting, we think of the $k$-th moment as the joint distribution of $k$ words in a document. 

The first moment of the topic model is given by the probability distribution over a single word, which corresponds to the average of the topic probability vectors.
\begin{equation}
	P(x_1) = \mathbb{E}_{\bm{t}_i \sim \bm{h}}[\mathbb{E}_{\bm{x}_1 \sim \bm{t}_i}[\bm{x}_1 | \bm{t}_i]] = \mathbb{E}_{\bm{t}_i \sim \bm{h}}[\bm{a}_i] = \sum_{i \in [k]} w_i \bm{a}_i
\end{equation}
The joint distribution of two different words in a document can then be expressed as the weighted sum of the means. Here we see that the exchangeability assumption makes our calculations a lot easier, since the noise is independent conditioned on the topic.
\begin{equation}
	\begin{split}
		P(x_1, x_2) & = \mathbb{E}_{\bm{t}_i \sim \bm{h}}[\mathbb{E}_{\bm{x}_1,\bm{x}_2 \sim \bm{t}_i}[\bm{x}_1, \bm{x}_2 | \bm{t}_i]] \\ & = \mathbb{E}_{\bm{t}_i \sim \bm{h}}[\mathbb{E}_{\bm{x}_1 \sim \bm{t}_i}[\bm{x}_1 | \bm{t}_i] \circledcirc \mathbb{E}_{\bm{x}_2 \sim \bm{t}_i}[\bm{x}_2 | \bm{t}_i]] \\ & = \mathbb{E}_{\bm{t}_i \sim \bm{h}}[\bm{a}_i \circledcirc \bm{a}_i] \\ & = \sum_{i \in [k]} w_i \bm{a}_i \circledcirc \bm{a}_i
	\end{split}
\end{equation}
We can now easily extend this concept to the co-occurrence of three words in a document.
\begin{equation}
	\begin{split}
		P(x_1, x_2, x_3) & = \mathbb{E}_{\bm{t}_i \sim \bm{h}}[\mathbb{E}_{\bm{x}_1,\bm{x}_2,\bm{x}_3 \sim \bm{t}_i}[\bm{x}_1, \bm{x}_2, \bm{x}_3 | \bm{t}_i]] \\ & = \mathbb{E}_{\bm{t}_i \sim \bm{h}}[\mathbb{E}_{\bm{x}_1 \sim \bm{t}_i}[\bm{x}_1 | \bm{t}_i] \circledcirc \mathbb{E}_{\bm{x}_2 \sim \bm{t}_i}[\bm{x}_2 | \bm{t}_i] \ \circledcirc \\ & \qquad \circledcirc \mathbb{E}_{\bm{x}_3 \sim \bm{t}_i}[\bm{x}_3 | \bm{t}_i]] \\ & = \mathbb{E}_{\bm{t}_i \sim \bm{h}}[\bm{a}_i \circledcirc \bm{a}_i \circledcirc \bm{a}_i] \\ & = \sum_{i \in [k]} w_i \bm{a}_i \circledcirc \bm{a}_i \circledcirc \bm{a}_i
	\end{split}
\end{equation}
The topic model's moments therefore directly correspond to the moments we presented in Equation \eqref{eq:moments}. \\[8px]

\noindent After discovering that we can represent the moments from both models through Equation \eqref{eq:moments}, we are interested in extracting the latent information from the moments. The problem we are facing now is that we are not able to acquire the $\bm{a}_i$s from the moments directly without simplifying assumptions. While we don't know the $\bm{a}_i$s at this point, we will introduce an orthogonality assumption on the $\bm{a}_i$s and relax this constraint again as we move on. 

Recall from Section \ref{sec:tpm}: if the $\bm{a}_i$s were orthogonal to each other, then the $\bm{a}_i$s would also directly correspond to the eigenvectors of the third-order moment tensor $\bm{\mathscr{M}}_3$. If we knew one specific $\bm{a}_i$, for example $\bm{a}_1$, then hitting $\bm{\mathscr{M}}_3$ with $\bm{a}_1$ on two dimensions yields the same eigenvector again, scaled by $w_1$:
\begin{equation}
	\label{eq:eigen_verify}
	\bm{\mathscr{M}}_3(\bm{I},\bm{a}_1,\bm{a}_1) = \sum_{i \in [k]} w_i \langle \bm{a}_i, \bm{a}_1 \rangle^2 \bm{a}_i = w_1 \bm{a}_1
\end{equation}
This directly corresponds to the concept of matrix eigenvectors. Recall: $\bm{M}\bm{v} = \bm{M}(\bm{I},\bm{v}) = \lambda\bm{v}$. Equation \eqref{eq:eigen_verify} therefore allows us to verify whether a certain vector corresponds to an eigenvector of $\bm{\mathscr{M}}_3$.

As we have discussed before, we can usually not assume that the $\bm{a}_i$s are orthogonal to each other. It is, however, possible to orthogonalize the third moment $\bm{\mathscr{M}}_3$, which implicitly also results in an orthogonalization of the $\bm{a}_i$s. This enables us to use the nice properties of Equation \eqref{eq:eigen_verify} on more general $\bm{a}_i$s.

\subsection{Orthogonalization Through Whitening}

Using the second moment $\bm{M}_2$, we can obtain a \textit{whitening transformation}\footnote{linear transformation that transforms a set of random variables with a known covariance matrix into a set of new variables whose covariance is the identity matrix (each variable has variance 1)}, that orthogonalizes $\bm{\mathscr{M}}_3$. Formally this is expressed as $\bm{W}^T\bm{M}_2\bm{W} = \bm{I}$, where $\bm{W}$ is called the whitening matrix. By applying $\bm{W}$ as a multilinear transformation on the third moment, we get
\begin{equation}
	\begin{split}
		\bm{\mathscr{V}} = \bm{\mathscr{M}}_3(\bm{W},\bm{W},\bm{W}) & = \sum_{i \in [k]} w_i (\bm{W}^T \bm{a}_i)^{\circledcirc 3}\\
		& = \sum_{i \in [k]} w_i \bm{v}_i \circledcirc \bm{v}_i \circledcirc  \bm{v}_i\\
	\end{split}
\end{equation}
where $(\bm{W}^T \bm{a}_i)^{\circledcirc 3} = (\bm{W}^T \bm{a}_i) \circledcirc (\bm{W}^T \bm{a}_i) \circledcirc (\bm{W}^T \bm{a}_i)$.

Since we are operating in a whitened space after the transformation of $\bm{\mathscr{M}}_3$, we should ensure that we can invert this transformation to recover the $\bm{A}$ in the original space. We are only able to perform this un-whitening if the $\bm{a}_i$s are linearly independent. Given the whitening matrix $\bm{W}$, we can relate the whitened latent factors (columns of $\bm{V}$) and the original latent factors (columns of $\bm{A}$) as follows: $\bm{V} = \bm{W}^T\bm{A}$.

Note that this whitening procedure leads to a dimensionality reduction, since $\bm{\mathscr{M}}_3 \in \mathbb{R}^{d \times d \times d}$ and $\bm{\mathscr{V}} \in \mathbb{R}^{k \times k \times k}$, which in turn imposes limitations on $k$ and $d$, namely $k \leq d$. While this is usually not a problem for topic models, since the size of the vocabulary can be assumed to be larger than the number of distinct topics, this imposes severe constraints on the GMM, where the number of components may easily exceed the dimensionality of the data. Hence, the presented method's applicability is limited for GMMs.

We can obtain the whitening matrix $\bm{W}$ through an eigendecomposition of the second moment $\bm{M}_2$ as follows:
\begin{equation}
\label{eq:whit_matr}
\boxed{
\begin{split}
	\bm{M}_2 = \bm{U}\text{Diag}(\tilde{\lambda})\bm{U}^T\ \Rightarrow \ \bm{W} & = \bm{U}\text{Diag}(\tilde{\lambda}^{-1/2})\\
	\bm{V} & = \bm{W}^T \bm{A}\text{Diag}(w^{1/2})
\end{split}
}
\end{equation}
The computation of $\bm{\mathscr{V}}$ for this specific transformation is therefore given by
\begin{equation}
	\begin{split}
		\bm{\mathscr{V}} = \bm{\mathscr{M}}_3(\bm{W},\bm{W},\bm{W}) & = \sum_{i \in [k]} \frac{1}{\sqrt{w_i}} (\bm{W}^T \bm{a}_i \sqrt{w_i})^{\circledcirc 3}\\
		& = \sum_{i \in [k]} \lambda_i \bm{v}_i \circledcirc \bm{v}_i \circledcirc  \bm{v}_i\\
	\end{split}
\end{equation}
After this transformation step, we are now capable of decomposing the tensor $\bm{\mathscr{V}}$ in order to uncover the latent structure present in the whitened third moment, which we will do through the tensor power method.

\subsection{Decomposition Through Tensor Power Method}
\label{sec:dec_tpm}

Recall that we can confirm whether $\bm{v}_1$ is an eigenvector of $\bm{\mathscr{V}}$ through the following tensor transformation:
\begin{equation}
	\bm{\mathscr{V}}(\bm{I},\bm{v}_1,\bm{v}_1) = \sum_{i \in [k]} \lambda_i \langle \bm{v}_i, \bm{v}_1 \rangle^2 \bm{v}_i = \lambda_1 \bm{v}_1
\end{equation}
What is still left is to find the $k$ dominant eigenvectors of the orthogonalized tensor $\bm{\mathscr{V}}$. We can achieve this by applying the tensor power method introduced in \ref{sec:tpm}. Note that, in order to correspond with the rest of the notation and variables introduced in this chapter, the deflation counter is referred to as $i$ and the power iteration counter as $j$. After randomly initializing a vector $\bm{v}_{i,j}$ we can feed the vector into the tensor power step and repeat this step until convergence (which generally only takes about a dozen iterations): 
\begin{equation}
	\label{eq:tensor_power_app}
	\bm{v}_{i,j + 1} = \frac{\bm{\mathscr{V}}_i(\bm{I},\bm{v}_{i,j},\bm{v}_{i,j})}{||\bm{\mathscr{V}}_i(\bm{I},\bm{v}_{i,j},\bm{v}_{i,j})||_2}
\end{equation}
Afterwards, we deflate the tensor, i.e. remove eigenvalue $\lambda_i$ with eigenvector $\bm{v}_i$ that we just extracted from the tensor as follows:
\begin{equation}
\label{eq:defl}
	\bm{\mathscr{V}}_{i + 1} = \bm{\mathscr{V}}_i - \lambda_i \bm{v}_i \circledcirc \bm{v}_i \circledcirc \bm{v}_i
\end{equation}
This process can be repeated until we have extracted $k$ dominant eigenvalue/-vector pairs from the tensor $\bm{\mathscr{V}}$.

As an alternative, one could also use other algorithms from the CPD family to solve this problem, as all of these methods try to recover a unique $\bm{A}$. Determining which algorithm is better suited depends on a number of different implementation decisions that we cannot address here in its entirety.

Finally, since we are not directly interested in the $\bm{v}_i$s but in the $\bm{a}_i$s, we still need to perform a backwards transformation through un-whitening via the following equation:
\begin{equation}
	\label{eq:a_est}
	\bm{A} = (\bm{W}^T)^\dagger \bm{V} \text{Diag}(\bm{\lambda})
\end{equation}
Note that $\bm{M}^{\dagger} = (\bm{M}^T\bm{M})^{-1}\bm{M}^T$ denotes the Moore-Penrose pseudo-inverse of the matrix $\bm{M}$.

\begin{algorithm}[!t!]
\begin{algorithmic}
\Procedure{tensor\_mixture\_est}{d, k}
	\State $\bm{X} \leftarrow$ read data / generate sample data using d, k
	\State Compute $1^{\text{st}}$ data moment (mean): $\bm{\mu} \leftarrow \frac{1}{\text{n}} \sum_{\text{i} = 1}^{\text{n}} \bm{x}_i$
	\State Compute $2^{\text{nd}}$ data moment (covariance): $\bm{\Sigma} \leftarrow \frac{1}{\text{n}} \sum_{\text{i} = 1}^{\text{n}} \bm{x}_i  \bm{x}_i^T$ 
	\State Decompose $2^{\text{nd}}$ data moment: $\bm{U}, \bm{S} \leftarrow \text{SVD}(\bm{\Sigma})$
	\State \textit{GMM only:} Extract shared variance: $\sigma^2_{\text{est}} \leftarrow \text{min}(\bm{S})$
	\State Compute whitening matrix $\bm{W}$ using Eq. \eqref{eq:whit_matr}
	\State Whiten the data: $\bm{X}_{whit} = \bm{X} \times \bm{W}$
	\For{i = 1,\ldots,k} 
		\State Generate random $\bm{v}_{\text{old}} \in \mathbb{R}^{k}$
		\State Normalize random vector: $\bm{v}_{\text{old}} \leftarrow \frac{\bm{v}_{\text{old}}}{||\bm{v}_{\text{old}}||}$
		\Repeat
			\State \small Compute multilin. transf.: $\bm{v}_{\text{new}} \leftarrow \frac{\bm{X}_{whit}^T \times (\bm{X}_{whit} \times \bm{v}_{\text{old}})^2}{\text{n}}$
            \If {$i > 1$} 
                \For{$l = 1,\ldots,i-1$}
                    \State Deflate tensor using Eq. \eqref{eq:defl}
                \EndFor          
            \EndIf
            \State $\bm{l} = ||\bm{v}_{\text{new}}||$
            \State Normalize new vector using Eq. \eqref{eq:tensor_power_app}
            \If{convergence criterion satisfied}
            	\State Add $\bm{v}_{\text{new}}$ as column $i$ to $\bm{V}_{\text{est}}$
            	\State Add $\bm{l}$ as entry $i$ to $\bm{\lambda}$
            \EndIf
			\State $\bm{v}_{\text{old}} \leftarrow \bm{v}_{\text{new}}$
		\Until{maximum number of iterations reached}
	\EndFor
	\State Perform back-transformation using Eq. \eqref{eq:a_est}
	\State \Return $\bm{A}_{\text{est}}\ (, \sigma^2_{\text{est}})$
\EndProcedure
\end{algorithmic}
\caption{Tensorized mixture model learning}
\label{alg:tensor_mixture}
\end{algorithm}

\subsection{Algorithm Summary}

To sum up the above approach to estimate latent variable models using tensor decomposition, we present the algorithm outline in Algorithm \ref{alg:tensor_mixture}.

For the more practically inclined reader we have also created both Matlab and Python scripts for estimating GMMs as part of this case study, which can be accessed here\footnote{\url{https://github.com/steverab/tensor-gmm}}. A similar Matlab code for estimating exchangeable hidden variable models can be found here\footnote{\url{https://bitbucket.org/kazizzad/tensor-power-method}}.

\section{Available software libraries}
\label{sec:lib_fram}

While lots of programming languages provide data structures for multi-dimensional arrays either as part of their standard libraries or as part of widely used external packages, we would briefly like to mention a few popular tensor libraries. These libraries usually provide a more optimized way of storing and treating tensors, as well as techniques for efficiently decomposing them using the algorithms we described in Section \ref{sec:tensor_decomp}. Since tensors and their decompositions have only started to gain traction in the applied computer science community over the recent years, most libraries are still only available for proprietary environments like Matlab. 

An overview of these libraries is given in Table \ref{tab:lib_fram}.

\setlength{\textfloatsep}{0.1cm}

\begin{table}
  	\begin{tabular}{ccc}
    	\toprule
    	Library&Available for&Source\\
    	\midrule
    	TensorLy & Python & \cite{tensorly}\\
    	N-way Toolbox & Matlab & \cite{n_way_toolbox}\\
    	pytensor & Python & \cite{pytensor}\\
    	scikit-tensor & Python & \cite{sktensor}\\
    	SPLATT & C/C++, Octave, Matlab & \cite{splatt}\\
    	rTensor & R & \cite{rtensor}\\
    	Tensor Toolbox & Matlab & \cite{tensor_toolbox}\\
    	Tensorlab & Matlab & \cite{tensorlab}\\
  		\bottomrule
	\end{tabular}
	\vspace{10px}
	\caption{Popular tensor libraries}
	\label{tab:lib_fram}
\end{table}
\section{Current research}

Finally, we would like to point out some of the current research directions at the intersection between tensors and machine learning. At the moment, most research is centered around the following two main questions \cite{ge:tensor}: \begin {enumerate*} [label=\itshape\alph*\upshape)]
\item \textit{How can we formulate other machine learning problems as tensor decompositions?} While some machine learning problems can already be solved very efficiently through tensor decompositions (see Section \ref{sec:tensor_ml}), the effort of determining whether tensor methods can also be beneficial to other machine learning algorithms, like neural networks \cite{anzamin:neuralnetworks}, is still ongoing.
\item \textit{How can we compute tensor decompositions under weaker assumptions?} While tensor decompositions usually have weak conditions for uniqueness, the requirements for effectively using them in machine learning settings are quite strong. Recall for example that the GMM estimation in Section \ref{sec:est_mm} requires $k \leq d$, which is a rather strong limitation.
\end {enumerate*}
\section{Conclusions}

Before closing with this paper, we would like to briefly recap some of the most important take-away messages. First, we have looked at the rotation problem for matrices, which prevented us from trying to find a unique low-rank decomposition for said matrix. Then, we have introduced basic tensor notation and properties and explained that low rank tensors are generally more rigid than low-rank matrices because of the interrelations between different slices along the tensor's dimensions. Afterwards, we introduced two of the most widely used tensor decomposition approaches, the CP decomposition and the Tucker decomposition, where we elaborated on how these decompositions are computed (which is often done through an ALS algorithm) and analyzed under which conditions these decompositions are unique. To build a bridge to the machine learning world, we have discussed how and why tensor decompositions are used in various machine learning sub-disciplines and also gave a detailed example of estimating Gaussian mixture models and simple topic models by using the method of moments and the tensor power method to extract the needed parameters. And lastly, we have given a list of the most prominent tensor libraries and a brief overview of current research questions in the tensor decomposition field.


\end{document}